%% file: main.tex
\documentclass[]{style/fairmeta}

\usepackage{hyperref}
\usepackage{url}
\usepackage{amsmath}
\usepackage{amssymb}
\usepackage{xargs}                      
\usepackage{graphicx}
\usepackage{svg}
\usepackage{tablefootnote}
\usepackage{subcaption}
\usepackage{wrapfig}
\usepackage{caption}
\usepackage{float}
\usepackage{changepage}

\title{
    Hardware Scaling Trends and Diminishing \\
    Returns in Large-Scale Distributed Training
}

\affiliation[1]{FAIR at Meta}
\affiliation[2]{Carnegie Mellon University}

\author[1,2,*]{Jared Fernandez}
\author[1]{Luca Wehrstedt}
\author[1]{Leonid Shamis}
\author[1]{Mostafa Elhoushi}
\author[1]{Kalyan Saladi}
\author[2]{Yonatan Bisk}
\author[2]{Emma Strubell}
\author[1]{Jacob Kahn}

\correspondence{Jared Fernandez at \email{jaredfern@cmu.edu}, Jacob Kahn at \email{jacobkahn@meta.com}}

\abstract{
  \input{sections/1_abstract.tex}
}
\begin{document}
\maketitle

\input{sections/2_intro.tex}
\input{sections/3_preliminaries.tex}

\input{sections/5_methods.tex}
\input{sections/6_analysis}
\input{sections/7_implications}
\input{sections/4_related_work.tex}
\input{sections/9_conclusion.tex}

\bibliography{references}
\bibliographystyle{style/iclr2025_conference}

\appendix
\input{sections/8_limitations.tex}
\input{sections/A_appx.tex}


\end{document}

%% file: sections/1_abstract.tex
    To train the exceedingly large neural networks required in modern applications, such as large language models (LLMs), model training is distributed across tens of thousands of hardware accelerators (e.g. GPUs), requiring orchestration of computation and communication across large computing clusters. In this work, we demonstrate that careful consideration of hardware configuration and parallelization strategy is critical for effective (i.e. compute- and cost-efficient) scaling of model training.
    We conduct an extensive empirical study of the performance of large-scale LLM training workloads across model size, hardware configurations, and distributed parallelization strategies with current best practices.
    In experiments with model sizes up to 70B parameters and utilizing up to 2048 H100 GPUs, we demonstrate that:
        (1) Naive scale out with Fully Sharded Data Parallelism (FSDP) incurs communication overhead which leads parallelization strategies previously thought to be sub-optimal in fact become preferable; and
        (2) scaling the total number of accelerators for training quickly yields diminishing returns even when hardware and parallelization strategies are properly optimized, implying poor marginal performance per additional unit of power or GPU-hour.

%% file: sections/2_intro.tex
\section{Introduction}
\label{sec:introduction}


The increasing size of state-of-the-art neural language models, which now contain in excess of hundreds of billions of parameters, yields larger computational workloads and memory requirements during training.  In this regime, the memory requirements from increasing numbers of model parameters and large batch sizes are such that the model parameters, activations, and optimizer states required for model training no longer fit within the memory of a single GPU accelerator. To address the memory limitations of a single device and to leverage the increased processing power of additional accelerators, the largest workloads necessitate distribution across thousands of hardware accelerators (i.e. GPUs and TPUs).

\begin{wrapfigure}{r}{0.45\textwidth}
    \centering
    \vspace{-1em}
    \includegraphics[width=\linewidth]{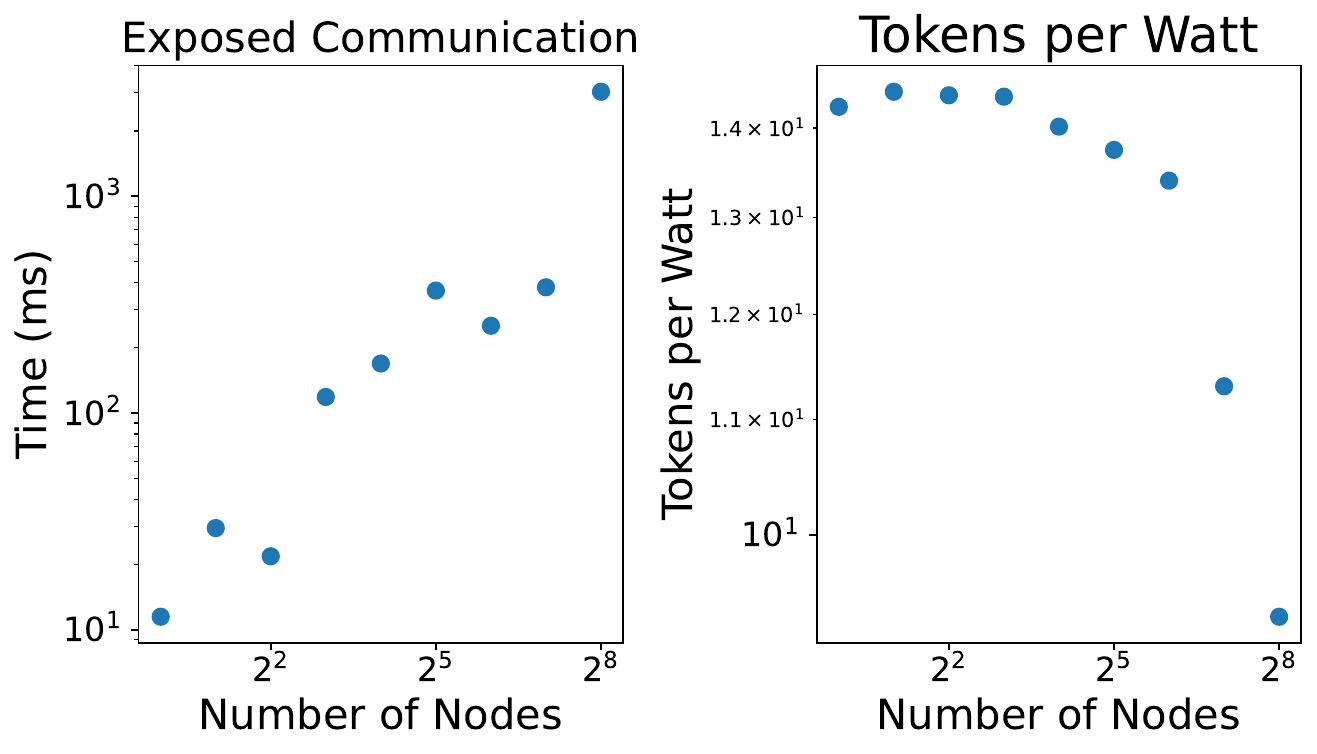}
    \caption[0.4\linewidth]{
    Despite minimal communication overhead on less than 32 nodes, increasing communication overhead leads FSDP to observe diminishing returns on power efficiency with over 30\% reduction at scale.
    }
    \vspace{-1em}
    \label{fig:figure-1}
\end{wrapfigure}

The need for training algorithms for distributing workloads across large numbers of accelerators has motivated the development of various data and model parallelism strategies -- discussed in more detail in \S\ref{sec:preliminaries} \citep{rasley2020deepspeed, Shoeybi2019MegatronLMTM, Zhao2023PyTorchFE, Li2020PyTorchD, ryabinin2023swarm}. Combining data, tensor and pipeline parallelism (3D parallelism) and sharded data parallelism (FSDP and DeepSpeed ZeRO) have been developed as primary methods to address memory limitations during training \citep{Shoeybi2019MegatronLMTM, shazeer2017outrageously, lepikhin2020gshard}. In particular, sharded data parallelism without model parallelism has emerged as one of the most common methods for langauge model training and been used in the training of open models such as: OLMo \citep{groeneveld2024olmo}, IBM Granite \citep{granite2024granite}, Apple OpenELM \citep{mehta2024openelm}, and Mosaic MPT \citep{MosaicML2023Introducing}.

Although many theoretical cost models have been developed to estimate the communication and computation performance of various parallelization methods \citep{qi2017paleo, cai2021tensoropt, pal2019optimizing, ModelBatchDomainParallelism, jia2019beyond}, existing approaches do not account for the full variety of components in modern training systems, including: model architecture, network topology, parallelisms, hardware speeds and architectures.  Previous work has empirically studied the performance and scaling properties of 3D parallelism \citep{narayanan2021efficient, hagemann2023efficient}, the scaling and efficiency properties of \textit{sharded parallelism strategies} and their interactions with model parallelism techniques are less well studied; despite its prevalence in practice (e.g. OLMo, Granite, OpenELM, MPT).



While there are stable distributed training recipes that perform well at large scale, the procedure for deciding on such configurations and their scaling properties are not well understood or documented; and the regimes in which selected parallelism strategies are communication and computation efficient is often unspecified.
Previous work \citep{narayanan2021efficient, hagemann2023efficient} has
studied the effects of various forms of model parallelism on training efficiency; we expand on this direction with studies across larger ranges of hardware and investigations of parallelism configurations not covered in previous studies. In particular, we consider the effects of Fully Sharded Data Parallelism (FSDP) on training efficiency and observe that its integration substantially impacts the choice of optimal training configurations. We show that prior work and existing best practices determined with model parallelism in the absence of FSDP, yield suboptimal performance and efficiency when combined with sharded data parallelism strategies. In addition, we conduct measurements of GPU power utilization and demonstrate that these existing approaches yield dramatically worse power efficiency, potentially worsening the energy and environmental cost of machine learning research and development \citep{strubell-etal-2019-energy,luccioni2024power,schwartz2020green}

In this work, we conduct an extensive empirical study across both parallelization strategies and hardware scales; and we
contribute the following:
\begin{adjustwidth}{-0.125cm}{}
\begin{itemize}
    \item A \textbf{large-scale empirical study} of distributed training across hardware setups, model sizes, and parallelism strategies, characterizing the scaling properties of sharded training; training on up to 2048 H100 GPUs in Section \ref{sec:analysis-dp} and \ref{sec:analysis-mp} and studying models up to 70B parameters in Section \ref{sec:analyis-model-size}
    \item Parallelization strategy recommendations which highlight that \textbf{model parallelism yields improved global throughput} despite prior work \citep{hagemann2023efficient, narayanan2021efficient} and conventional knowledge suggesting that model parallelism lowers hardware utilization.
    \item \textbf{Analysis of real-world cost metrics} showing that total GPU power draw and available FLOPS scale linearly with the number of devices, despite diminishing returns in throughput; resulting in reduced power efficiency and lower hardware utilization with greater parallelization (see Figure \ref{fig:figure-1}).
    \item \textbf{Comparisons across GPU hardware generations} suggesting that future improvements in computational throughput will only marginally improve overall throughput and power efficiency absent network fabric advancements and increased accelerator memory capacity in Section \ref{sec:analysis-hw-speed}.
\end{itemize}
\end{adjustwidth}

%% file: sections/3_preliminaries.tex
\section{Preliminaries} 
\label{sec:preliminaries}

In this section, we review commonly used parallelism techniques used in distributed training of large neural networks. The primary goals of distributed training are to: (1) enable model training with batch sizes and parameters that exceed the memory of individual GPUs; and (2) leverage the parallel processing power of additional hardware accelerators.

\subsection{Parallelization Strategies}
\label{sec:distributed-algos}
Below, we provide a brief taxonomy of commonly used distributed training algorithms and memory optimizations. In practice, these algorithms are not mutually exclusive and are often combined.

\paragraph{Data parallelism} \citep{dean2012large} replicates model parameters and optimizer states across GPUs with each device operating over a subset of examples in the global minibatch. After performing local forward and backward passes on their allocated minibatches, GPUs exchange and accumulate their partial gradients via an \texttt{AllReduce} collective such that each device obtains an identical global gradient and ensuring model update. Data parallelism exhibits favorable communication properties as the \texttt{AllReduce} operation is non-blocking.

\paragraph{Sharded Data Parallelism} alleviates the memory requirements of vanilla data parallelism by sharding model parameters, optimizer states, and gradients across devices in a data parallel group; rematerializing and updating necessary weights on-the-fly via \texttt{AllGather} and \texttt{ReduceScatter} operations. Fully-Sharded Data Parallelism \cite{Zhao2023PyTorchFE} and DeepSpeed ZeRO \cite{rasley2020deepspeed,rajbhandari2020zero} are commonly used sharded data parallelism strategies which enable training of large models without model parallelism. In contrast to distributed data parallel, sharded data parallelism introduces blocking communication operations to perform \texttt{AllGather} of model parameters; some of which can be overlapped by prefetch of subsequent layers during the previous layer computations.

\begin{figure*}
\vspace{-2em}
\centering
    \begin{subfigure}{.48\textwidth}
      \centering
      \includegraphics[width=\linewidth]{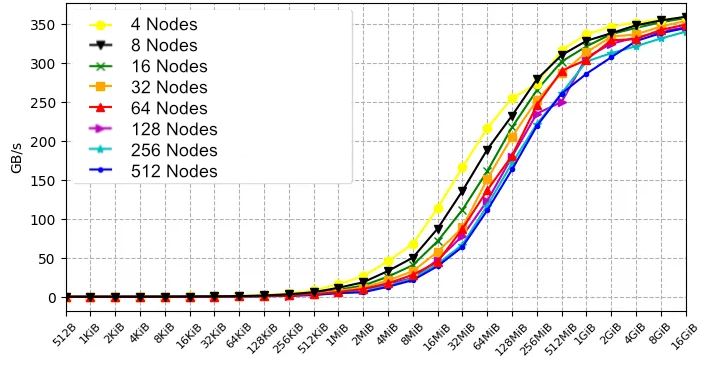}
      \caption{
        \footnotesize
        Bandwidth of NCCL \texttt{AllReduce} using a tree algorithm and scales well with number of nodes (i.e. higher bandwidth).
        }
      \label{fig:allreduce_microbenchmarks}
    \end{subfigure}
    \hfill
    \begin{subfigure}{.48\textwidth}
      \centering
      \includegraphics[width=\linewidth]{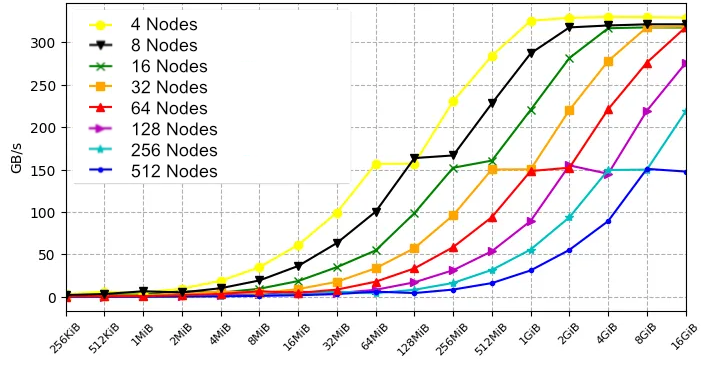}
      \caption{
        \footnotesize
        Bandwidth of NCCL \texttt{AllGather} using ring algorithms; scales poorly with the number of nodes (i.e. lower bandwidth).
      }
      \label{fig:allgather_microbenchmarks}
    \end{subfigure}
\caption{
    Bandwidth measurements in GB per second of NCCL primitives on DGX H100 servers with eight GPUs per node, connected with InfiniBand, across world sizes from 4 to 512 nodes.
}
\vspace{-2em}
\label{fig:nccl_microbenchmarks}
\end{figure*}

\paragraph{Model parallelism} shards model parameters across GPUs; each shard operates on the same minibatch simultaneously. In this setting, activations and their respective gradients are sent across GPUs.

\begin{adjustwidth}{-0.125cm}{}
\begin{itemize}
  \item \textbf{Tensor Parallelism} \citep{Shoeybi2019MegatronLMTM, shazeer2018meshtensorflowdeeplearningsupercomputers, zheng2022alpa} shards model parameters along hidden dimensions across devices such that each GPU computes a partial sum of the intermediate activations, which are then aggregated across the tensor parallel group via an \texttt{AllReduce}.As the full set of activations are required for the subsequent layer, Tensor Parallelism introduces \textit{blocking communication} for synchronization of intermediate activations across model parallel groups.
  \item \textbf{Pipeline Parallelism} \citep{Huang2018GPipeET, harlap2018pipedreamfastefficientpipeline, li2021chimera, li2021terapipe} shards model depthwise along with groups of layers being partitioned and allocated across devices; activations are then forwarded between devices via point-to-point communications. For all devices to be active at once, an input minibatch is split into microbatches which are staggered and pipelined according to various schedules \citep{Narayanan2019PipeDreamGP,lamy2023breadth}.  ``Pipeline bubble’’ \citep{hennessy2017computer}, in which devices remain idle while awaiting data or instructions from other stages reduces the efficiency of pipelining.
\end{itemize}
\end{adjustwidth}

Sequence \citep{li2023sequence, korthikanti2023reducing, jacobs2023deepspeed} and context parallelism  \citep{liuringattention, yang2024context} are techniques for reducing memory requirements for intermediate activations by partitioning and sharding along the sequence dimension.
The joint combination of data, tensor, and pipeline parallelism techniques is frequently utilized in what is known as 3D parallelism to achieve higher communication efficiency \citep{Shoeybi2019MegatronLMTM}; or 4D parallelism when utilizing these methods along with context parallelism when training with longer sequence lengths \citep{dubey2024llama} .

\paragraph{Communication-Computation Overlap} ~
Moving data over networks between accelerators utilizes distinct GPU resources unrelated to computation (e.g., dedicated copy engines, NVLink/NVSwitch) and can execute in parallel with computation. Overlapping communication and computation maximizes distributed training efficiency -- it facilitates hiding communication latency, leading to near-perfect scaling. \textit{Exposed communication}, that is communication that is executed without simultaneous computation leaves GPU's compute resources under-utilized.


\subsection{Communication Primitives and Libraries}
\label{sec:communication-in-deep-learning-systems}
Modern deep learning frameworks \citep{paszke2019pytorch,Abadi_TensorFlow_Large-scale_machine_2015,jax2018github} leverage specialized collective communications libraries, such as NCCL,
RCCL,
or XLA.
In particular, NCCL is used as the backend primary communication library for distributed operations across Nvidia GPUs.

In Figure \ref{fig:nccl_microbenchmarks}, we empirically benchmark the \texttt{AllReduce} and \texttt{AllGather} operations performance with the NCCL library. The \texttt{AllReduce} collective used in vanilla distributed data parallelism and tensor parallelism to aggregate parameter gradients and intermediate activations, respectively -- is supported by both Tree and Ring based algorithms in NCCL and observes favorable scaling properties as nodes increases. In contrast, \texttt{AllGather} and \texttt{ReduceScatter} are used for parameter rematerialization and gradient updates by FSDP and ZeRO; is only supported by Ring algorithms in NCCL, and quickly becomes latency-bound as the number of devices increases.

%% file: sections/5_methods.tex
\section{Experimental Methodology}
\label{sec:experimental_methodology}
In the following sections, we investigate the effects of scaling training workloads on end-to-end system performance and communication and computation volume. In particular, we conduct experiments across: distributed parallelization strategies, numbers of accelerators, hardware generation, model sizes, and input shapes (i.e. context length). Additional details on hardware and framework configurations are provided in Appendix \ref{appx:exp_details}.

\paragraph*{Model Architectures}
We conduct our experiments with the Llama-2 decoder-only transformer \citep{dubey2024llama, touvron2023llama} as a representative large language model.  We utilize the AdamW optimizer \citep{loshchilov2018decoupled, KingBa15} and train on examples with a context length of 4096 and tokenized with a vocabulary of 32K; with data sampled from Wikipedia and StackExchange.

\textbf{Hardware Configuration} ~
We evaluate distributed training on datacenter clusters containing 8-GPU NVIDIA DGX nodes from the Ampere (80GB A100) and Hopper (80GB H100) architectures, with additional experiments on Volta GPUs (32 GB V100) in Appendix \ref{fig:v100}. We conduct our primary experiments on hardware scales between 1 and 32 eight-GPU nodes, with additional experiments up to 256 nodes, or 2,048 GPUs -- to simulate pretraining scales.

\textbf{Parallelization Strategies} ~
We examine data, tensor, and pipeline parallelization strategies (colloquially known as 3D parallelism as described by \cite{Shoeybi2019MegatronLMTM, rasley2020deepspeed} and used in \cite{dubey2024llama, bigscience_workshop_2022}). Models are trained with Fully-Sharded Data Parallelism with explicit prefetching and without parameter resharding during the forward pass (i.e. FSDP, \cite{Zhao2023PyTorchFE}) as in Llama-3.1 training equivalent to DeepSpeed ZeRO Stage 2.

We examine a range of group sizes for tensor and pipeline parallel strategies for, as described in Section \ref{sec:preliminaries}, ranging from group sizes of 1 (i.e. single GPU training with no parallelization) up to group sizes of 16 (i.e. requiring parallelism groups across multiple nodes).

\paragraph*{Performance Metrics}
\label{sec:measuring-performance}
To understand the effects of both hardware and model scaling on end-to-end global and local per-device performance hardware utilization, we examine the following performance and efficiency indicators:

\begin{adjustwidth}{-0.125cm}{}
\begin{itemize}
    \item \textbf{Throughput} is the rate at which examples are processed. We compute the estimated per-device \textit{words per second} (WPS) and the global words per second across all devices.
    \item \textbf{Computational and communication load} is measured as the total execution time for CUDA and NCCL kernels, respectively. We calculate the total computation and communication load by aggregating CUDA and NCCL kernels from PyTorch execution traces.
    \item \textbf{Communication efficiency} is measured as the time in which communication kernels are exposed or overlapped with concurrent computation.
    \item \textbf{Hardware utilization} is measured as the number of floating point operations per second (FLOPS); alternatively, as Model FLOPS Utilization (MFU, \cite{chowdhery2023palm}) which is the observed FLOPS as a percentage of the hardware's reported theoretical maximum.
    \item \textbf{Power utilization} is reported as the per-GPU power draw measured as the the average power draw with NVML\footnote{https://developer.nvidia.com/management-library-nvml}.
\end{itemize}
\end{adjustwidth}

Metrics are aggregated from 60 training iterations, discarding the first 10 iterations to allow for stabilization of performance during the initial training iterations. Reported metrics are aggregated for the last 50 iterations.



%% file: sections/6_analysis.tex
\section{Performance Analysis}
\label{sec:effects-of-scaling}

\subsection{Weak Scaling: Variable Global Batch Size}
\label{sec:analysis-dp}

\begin{figure*}[ht]
    \centering
    \vspace{-1em}
    \includegraphics[width=\textwidth]{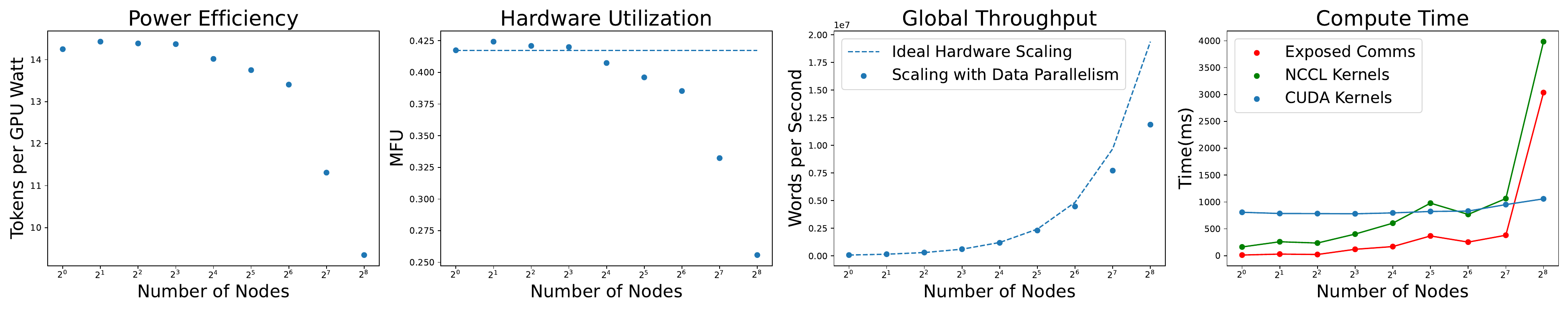}
    \vspace{-1.25em}
    \caption{
        \small
        In FSDP training of Llama-7B, scaling the number of nodes and data parallel replicas \textit{reduces hardware utilization and power efficiency} due to increasing exposed communication derived from increases in the size of communication kernels relative to fixed size computation kernels. Global throughput observes sub-linear scaling despite approximately linear increases in the total power utilization with number of nodes.  ``Ideal Hardware Scaling'' corresponds to  expected throughput assuming  additional accelerators yield linear increases in throughput.
    }
    \label{fig:world_size_scaling}
\end{figure*} 

\begin{wrapfigure}{r}{0.35\textwidth}
    \centering
    \vspace{-2em}
    \includegraphics[width=0.325\textwidth]{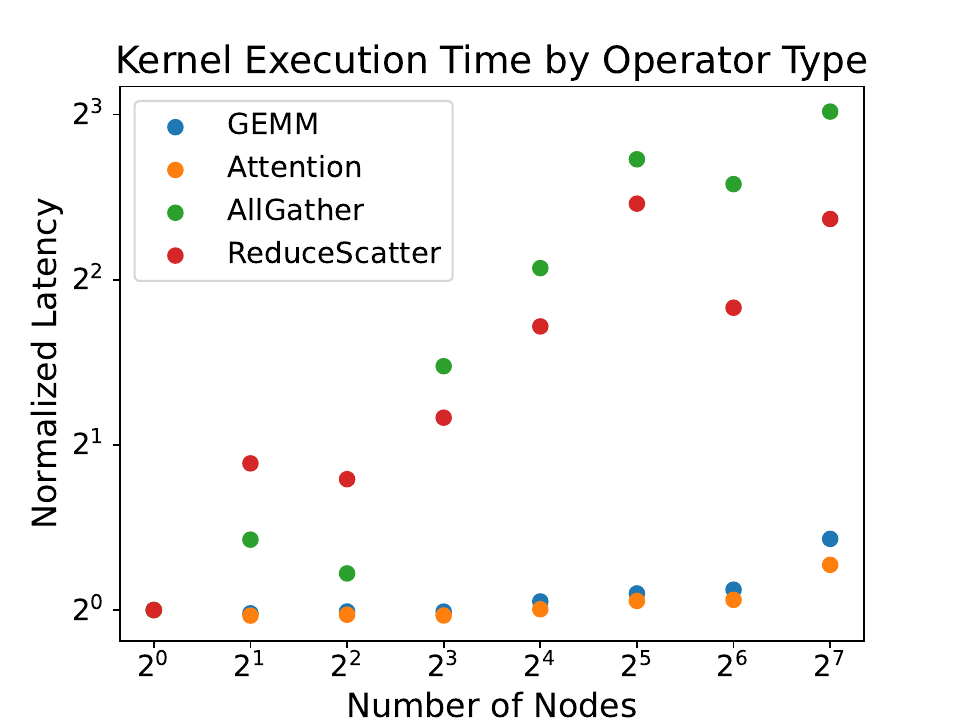}
    \caption{The relative execution time of both \texttt{AllGather} and \texttt{ReduceScatter} collectives scale with hardware world size.}
    \label{fig:kernel_scaling}
    \vspace{-1em}
\end{wrapfigure}

We first consider a \textit{weak scaling} setting in which the \textit{per-device workload} is kept constant as the number of GPU accelerators is increased. Each device carries a data parallel replica of a Llama 7B with a local batch size of 2 examples, and is trained with FSDP without any model parallelism. This is representative of training settings in which there are insufficient devices to train a model without gradient accumulation; and increasing the number of devices can be used to reduce the number of gradient accumulation steps. 

In Figure \ref{fig:world_size_scaling}, we examine the effects of weak scaling of data parallel training instances across increasing numbers of accelerators from 8 GPUs up to 2048 GPUs.  As expected, increasing the number of devices yields increases in global throughput as global batch size increases (i.e. Gustafson's Law for weak scaling; \cite{gustafson1988reevaluating}). At small scales (i.e. when training using a limited number of devices), the cost of collective communication kernels is low relative to the cost of computation -- and the communication overhead of weak scaling is minimal as non-blocking communication from FSDP can be hidden by executing data transfer and computation operations concurrently.

However, as discussed in Section \ref{sec:preliminaries}, increasing the degree of sharded data parallelism incurs larger collective communication costs for materialization of parameters via \texttt{AllGather} during the forward pass and gradient updates during the backward pass via \texttt{ReduceScatter}; with the latency of both operations scaling with number of nodes as observed in Figure \ref{fig:kernel_scaling}. As a result, the total execution time for NCCL communication kernels and volume of exposed communication scales with the number of compute nodes limiting the extent to which weak scaling can be applied to distributed sharded data parallel training --  matching the expected behavior observed for the communication collectives in Figure \ref{fig:allgather_microbenchmarks}.

While the communication volume scales with node count, the per-device CUDA computation kernel execution time remains constant and becomes dominated by communication. As a result, exposed communication is unavoidable at scales \textit{larger than 128 GPUs} and the hardware utilization decreases as there is insufficient computation to saturate the GPUs while waiting for the execution of larger communication kernels -- this results in reductions the marginal speedup of global throughput and decreased local throughput as the number of devices increases.


While the per-device throughput scales sublinearly with the number of devices, the total power utilization scales approximately linearly which results in substantially worse real-world efficiency in GPU-hours and energy impact (i.e. fewer tokens processed per watt). When scaling from 128 to 2048 GPUs, the observed TFLOPS and words-per-second throughput decrease by 37.22\% due to increasing exposed communication. Although the accelerator is largely idle on large scales and operates at lower arithmetic intensity, the power draw per GPU is roughly constant, only decreasing by 5.87\% from 658W to 620W. As a result, the \textit{overall power efficiency of the system likewise decreases with hardware scale} as seen in Figure \ref{fig:world_size_scaling}.


\subsection{Strong Scaling: Fixed Global Batch Size}
\label{sec:analysis-worldsize}

\begin{figure*}[h]
    \centering
    \includegraphics[width=\textwidth]{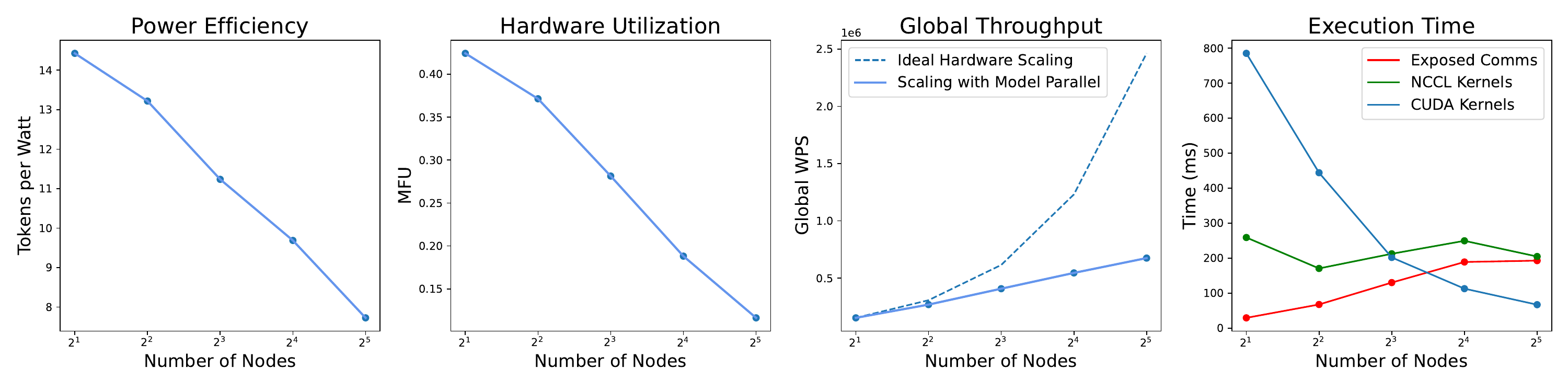}
    \vspace{-2em}
    \caption{
        \small
        \textit{Training with Fixed Global Batch Size Over Increasing Number of Nodes}. We select the optimal parallelization strategy as determined by the experimental results displayed in Figure \ref{fig:mp-pp} for configurations of up to 32 H100 nodes to train with global batch size of 32. Even with optimal parallelization strategies, local throughput and hardware utilization declines with world size.
    }
    \vspace{-1em}
    \label{fig:fixed_bs}
\end{figure*}

 We now examine the effects of strong scaling the number of accelerators to train workloads with a \textit{fixed global batch size}, which results in decreasing effective local per-device batch sizes as the number of devices increases. This is representative of industry settings where excess compute resources can be allocated for a single training run; and there is a desire to minimize the time to complete a training run as opposed to maximizing the hardware utilization. 

In Figure \ref{fig:fixed_bs}, we show that when training with a fixed global batch size of 32 examples across 2 to 32 nodes -- allocation of additional devices yields diminishing returns in global throughput and reduced local hardware utilization and power efficiency. To distribute a fixed workload across more devices, it is necessary to introduce excess degrees of model parallelism which results in insufficient amounts of computation being allocated to each accelerator; which we observe as reduced execution time for CUDA kernels. At sufficiently large scales, excess parallelism causes previously compute-bound workloads to become communication bound and yields reductions in hardware utilization, which we observe in decreases in MFU from  $40\%$ when training with 2 nodes to less than $15\%$ with 32 nodes. Practically, the overheads of strong scaling are especially apparent when using more than 4 nodes or 32 GPUs, as per-device workload sizes decrease and model parallelism becomes necessary. 

In Appendix~\ref{appx:fixed_global_batch_size}, we conduct additional strong scaling experiments at full pretraining scale training both LLAMA-7B and 70B models on between 512 to 2048 GPUs, with limited marginal returns for increasing the number of hardware accelerators and observe decreases in MFU local hardware utilization by more than $30\%$.

\subsection{Scaling Model Parallelism}
\label{sec:analysis-mp}
\begin{figure*}
    \centering
    \includegraphics[width=\textwidth]{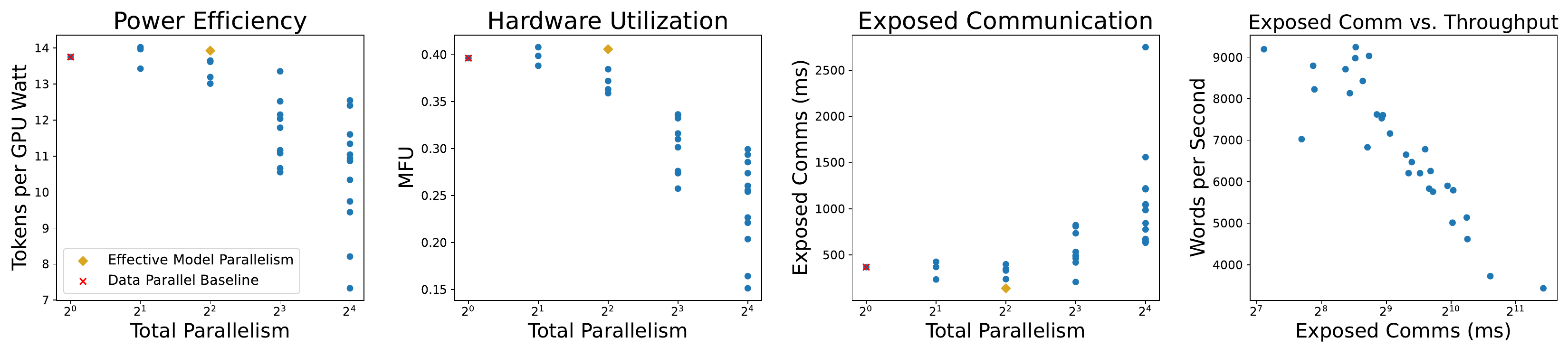 }
    \caption{
        \small
        \textit{Model parallelism increases FSDP throughput.}
        In model parallel training of Llama-7B with a fixed global batch size (512) and fixed number of accelerators (256 GPUs), there exist model parallel strategies
        that \textit{increase} training throughput,  hardware utilization, and power efficiency by reducing the total exposed communication. 
    }
    \vspace{-1.5em}
    \label{fig:mp-pp}
\end{figure*}

As observed with both strong and weak scaling, fully sharded data parallel training of large neural networks suffers from communication bottlenecks when conducted over sufficiently parallel hardware platforms due to increasing costs of \texttt{AllGather} and \texttt{ReduceScatter} at scale.

Model parallelism is commonly used to complement data parallel training and reduce the memory requirements of a training workload to fit within the memory of each individual device. Additionally, model parallelism enjoys another beneficial property in which it can reduce the sizes of collective communication operations; as separate data parallel replicas are maintained for each model parallel group (i.e. data parallel collectives are executed over world sizes of $\frac{\text{Number of Devices}}{\text{Total Degree of Model Parallelism}}$, rather than over the Total Number of Devices) -- where Total Degree of Model Parallelism is the product of Tensor and Pipeline parallelism group sizes.

In Figure~\ref{fig:mp-pp}, we search viable parallelism strategies for Llama 7B on 32 nodes with an effective local batch size of two and observe that small degrees of total model parallelism (i.e. tensor or pipeline parallel degrees of 2 or 4) reduce the amount of \textit{exposed communication} and \textit{increase throughput}. Although both tensor and pipeline parallelism introduce additional communication operations, both techniques reduce the data parallel group sizes of the FSDP \texttt{AllGather} and \texttt{ReduceScatter} collectives; yielding higher throughput, hardware utilization, and power efficiency.

Furthermore, in Figure \ref{fig:mp_across_generations}, we find that both tensor and pipeline parallelism are effective in reducing exposed communication; yielding higher words-per-second relative to the data parallel baseline. When scaling to more devices, we observe that the size of collective communications grows which necessitates increasing degrees of model parallelism to reduce the size of FSDP collective communications in Figure \ref{fig:al}.

Notably, there is a limit to the extent to which model parallelism reduces exposed communication and improves throughput -- as the \texttt{AllReduce} kernels required for Tensor Parallelism and bubbles introduced by pipeline parallelism grow with the degree of model parallelism. These communication costs become especially large when the parallelism occurs over multiple nodes as it relies on slower internode fabric (e.g. InfiniBand) -- as noted in Figure \ref{fig:mp_across_generations}, where there is substantial increases in exposed communication for tensor and pipeline parallelism strategies which are sharded at larger than 8 devices (i.e. across multiple nodes).

\subsection{Scaling Hardware Speeds}
\label{sec:analysis-hw-speed}

\begin{wrapfigure}{r}{0.4\textwidth}
    \centering
    \vspace{-4em}
    \begin{subfigure}[t]{0.195\textwidth}
        \includegraphics[width=1\textwidth]{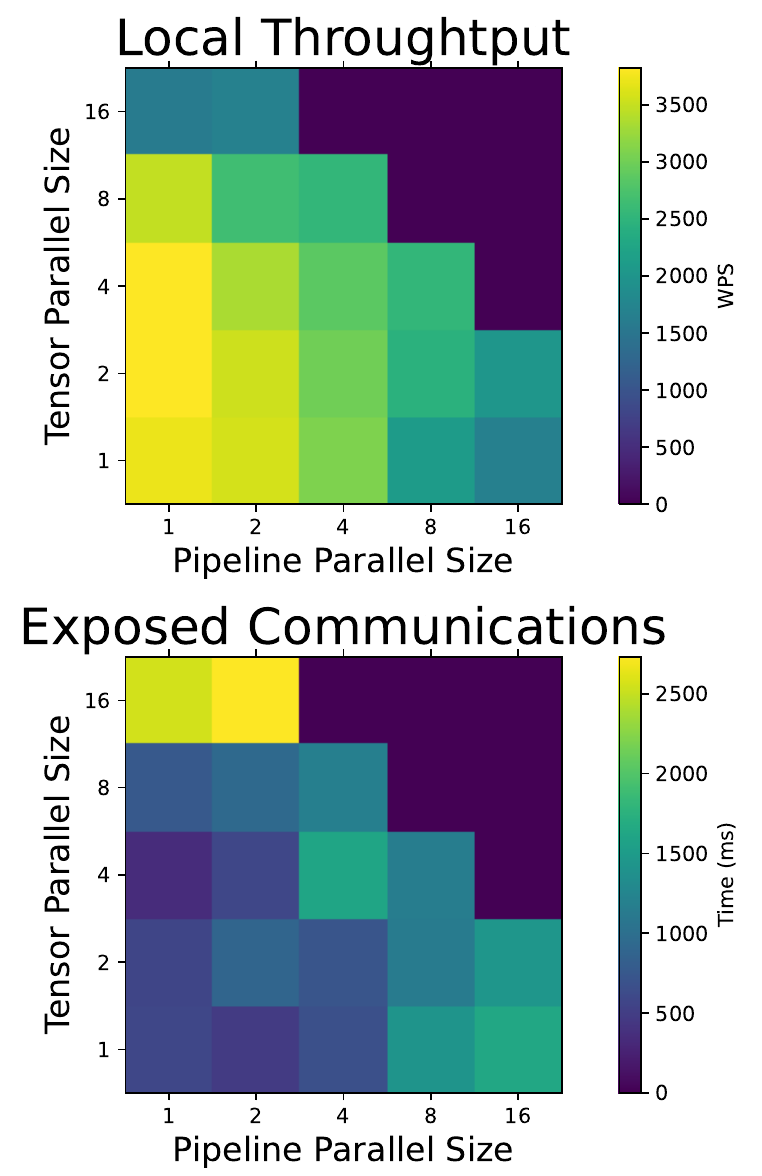}
        \caption{32 Nodes A100}
    \end{subfigure}
    \begin{subfigure}[t]{0.1925\textwidth}
        \includegraphics[width=1\textwidth]{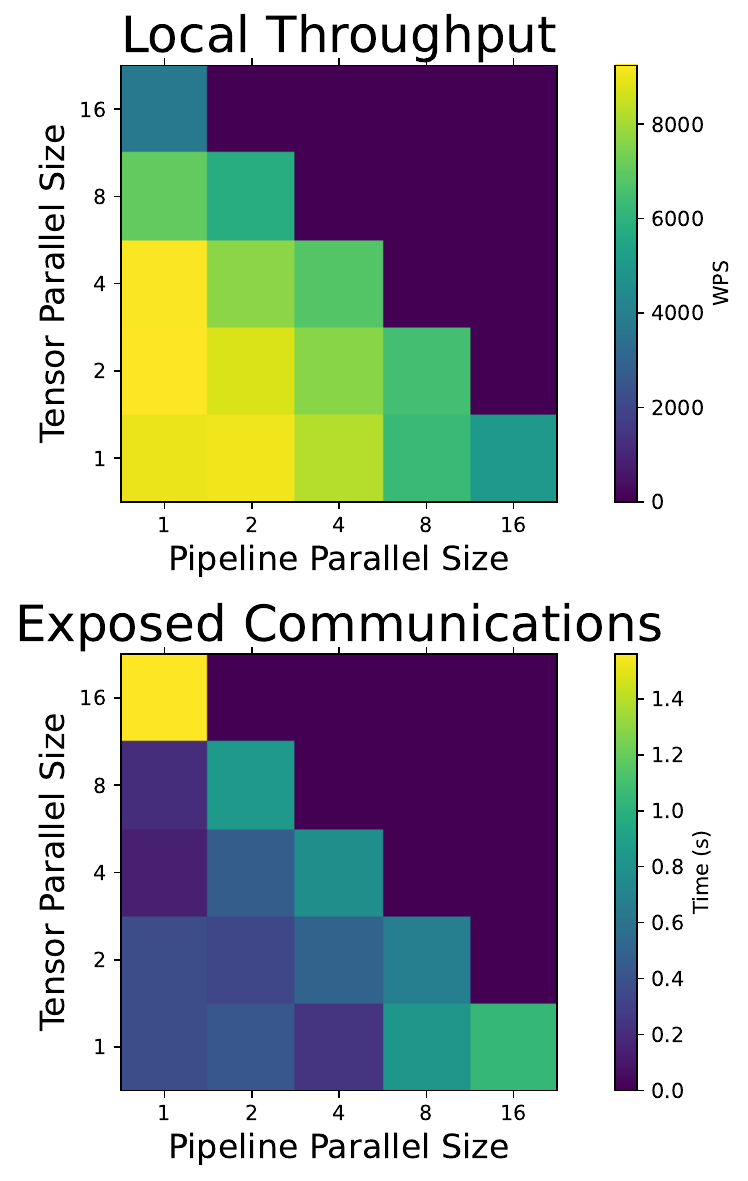}
        \caption{32 Nodes H100}
    \end{subfigure}
    \vspace{-1em}
    \caption{
        \small
        \textit{Model Parallelism Improves Throughput.} Increasing degree of either tensor and pipeline parallelism yields improved throughput and less exposed communications compared to data parallel baselines.
    }
    \vspace{-4em}
    \label{fig:mp_across_generations}
\end{wrapfigure}

In Figure \ref{fig:mp_across_generations}, we examine the effects of scaling the hardware speed with comparisons between DGX-A100 and H100 clusters. In both cases, there exist model parallelism configurations which both increase the overall throughput and reduce the amount of exposed communication relative to data parallel baselines (i.e. total model parallelism of one).

When comparing the training performance of previous generation A100 to faster H100 hardware, with the optimal parallelization strategy for each platform, the MFU hardware utilization \textit{decreases} from $59.67\%$ to $40.77\%$
The reduction in hardware utilization can be attributed to increases in exposed communication (+$12.83\%$) that emerge due to asymmetric improvements in communication and computation speeds (i.e. \texttt{bf16 FLOPS} more than triples whereas NVLink and HBM bandwidth increase by ~50\%; See  Table \ref{tab:gpu_specs}). 

Between the A100 and H100 architectures, the extent to which training is \textit{communication bound increases further with hardware generation}. Improvements to computation speed that outpace increases in data transfer speeds, result in computational kernels executing more quickly which make overlap with communication difficult (See Table \ref{tab:gpu_specs}).  
In Appendix \ref{appx:v100}, we conduct additional experiments with V100 GPUs in which we confirm that the highest throughput is achieved with model parallelism.

\subsection{Scaling Size of Model Architecture}
\label{sec:analyis-model-size}
 \begin{wrapfigure}{r}{0.40\textwidth}
 \vspace{-4em}
  \begin{center}
    \includegraphics[width=0.40\textwidth]{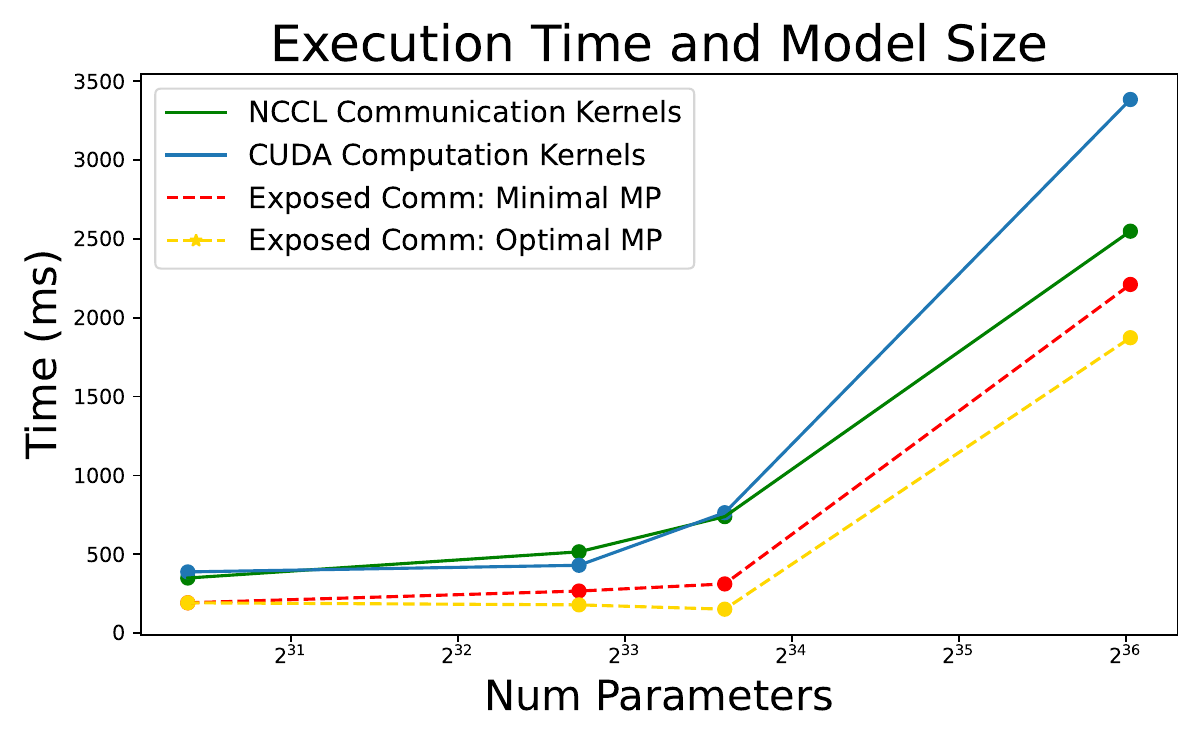}
  \end{center}
  \caption{ \small
        \textit{Communication \& Computation Both Scale with Model Size}. As computation load increases with model size, so does total and exposed communication. At all model scales, model parallelism reduces exposed communication.
    }
    \vspace{-2em}
    \label{fig:model_scaling}
\end{wrapfigure}

We examine the effects of scaling the size of the neural network architectures across 1B, 7B, 13B, and 70B parameters. One might assume that increases in model parameterization solely increases the size of computation while leaving communication unaffected. However, as the number of parameters in a model scale, the volume of communication required for parameter materialization and gradient scattering increases jointly with the size of the computational operations (i.e. matrix operations with larger hidden dimensions). In Figure \ref{fig:model_scaling}, we consider the optimal model parallelism strategy for each model architecture by sweeping viable tensor and pipeline parallel configurations and observe that the volume of \textit{exposed communication} likewise increases with model size, resulting in lower hardware utilization as models scale.

Additionally, we find that across architecture scales, there exist model parallelism strategies beyond the data parallel baseline or the minimal degree of model parallelism (for the 70B parameter model) that reduce the volume of exposed communication for all model sizes; and yield higher hardware utilization and throughput.
\subsection{Scaling Context Length}

\begin{figure*}[h]
    \vspace{-1em}
    \begin{subfigure}[t]{\textwidth}
        \centering
        \includegraphics[width=0.9\textwidth]{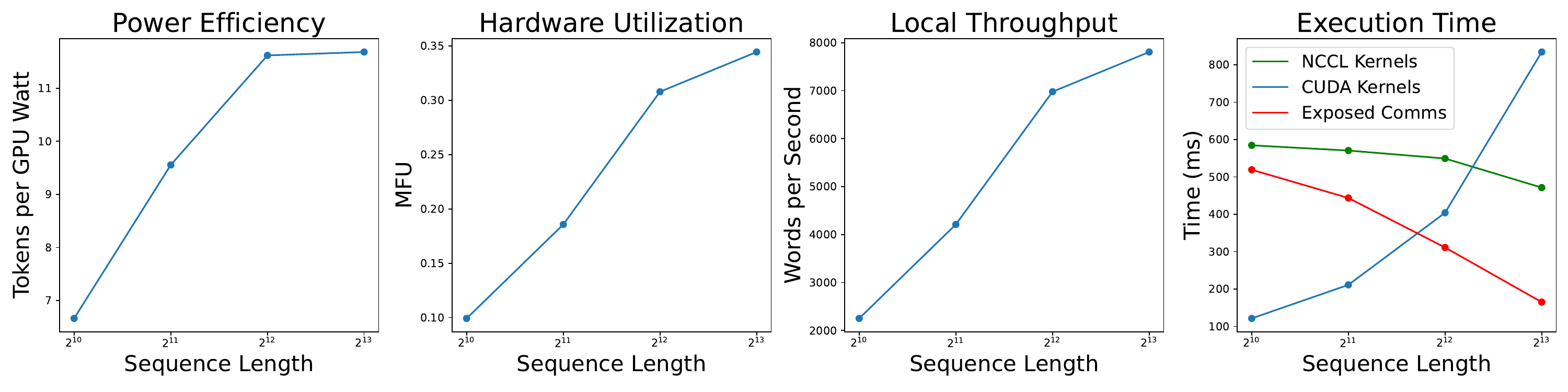}
    \end{subfigure}
    \caption{
        Increased sequence lengths yields larger compute kernels which better overlap with NCCL communication kernels, resulting in \textit{lower exposed communication, higher hardware utilization and power efficiency}.
    }
    \vspace{-1em}
    \label{fig:compute_scaling}
\end{figure*}

Finally, we examine the effects of varying the context length in Figure \ref{fig:compute_scaling}. When GPU memory is available, increasing the sequence length increases the computational workload allocated to each device without increasing the communication load, yielding improved the throughput, hardware utilization and power efficiency. However, reparameterization of the training process in this manner is often infeasible, as alterations to per-batch sequence length affects training dynamics predicted by computation-architecture scaling laws \citep{kaplan2020scaling, hoffmann2022empirical}.

%% file: sections/7_implications.tex
\section{Recommendations and Implications}
\vspace{-0.5em}
\label{sec:trends-and-takeaway}
We summarize our findings with directions for future work and best practices for researchers and practitioners.

\paragraph{Model parallelism alleviates growing communication costs of FSDP.} Prior work \citep{hagemann2023efficient, Shoeybi2019MegatronLMTM} studying 3D parallelism for large-scale training is conducted without the use of FSDP or ZeRO, and concludes that data parallel training is generally preferred to model parallelism when models fit within device memory. However, the collective communication primitives required by FSDP and ZeRO exhibit worse scaling properties than those used by standard distributed data parallel, as seen in Figures \ref{fig:nccl_microbenchmarks}, \ref{fig:world_size_scaling}, \ref{fig:fixed_bs}. We demonstrate that the increasing communication overhead of FSDP at scale can be mitigated by both tensor and pipeline parallelism. 

In particular, we observe that standard FSDP training of 7B LLMs becomes unavoidably communication bound when training on more than 128 H100 GPUs. Beyond this scale, tensor parallelism at degrees of 2 or 4 achieves better or comparable throughput to the FSDP baseline. In our largest experiments at 2048 GPUs, introducing tensor parallelism yields a +52.60\% increase in WPS throughput while only drawing 30W more in average GPU power per-device.

\paragraph{Power efficiency and  hardware utilization exhibit diminishing returns at scale.} As number of devices scales, energy efficiency decreases because the per-device computational throughput (FLOPS) decreases, despite power utilization remaining roughly constant (See Figure \ref{fig:world_size_scaling} and \ref{fig:fixed_bs}). Inefficient scaling and parallelization methods will worsen the energy efficiency and environmental cost of model training, as architectures and hardware platforms grow in scale \citep{strubell-etal-2019-energy, luccioni2023countingcarbonsurveyfactors, luccioni2024power, schwartz2020green}. Rather than relying on synchronous training with a single large model, research in alternative training formulations that reduce communication overhead are required to improve model efficiency as models scale; such as via federated averaging, asynchronous training, ensembles and modular model architectures. 

\paragraph{Asymmetric improvements in hardware increase communication boundedness.} 
Hardware improvements have resulted in disproportionate growths in compute speeds that have outpaced improvements to memory and network speeds. As a result, model training is increasingly communication bound with the identical training workload observing a nearly 20\% decrease in hardware utilization when using H100, as compared to A100 hardware (Section \ref{sec:analysis-hw-speed}).  

When training at large scales, faster interconnects are needed in addition to improvements in accelerator speed.  Likewise increasing node size, such as with NVIDIA's GB-200 \footnote{\hyperlink{https://resources.nvidia.com/en-us-dgx-systems/dgx-superpod-gb200-datasheet}{NVIDIA GB-200 Datasheet}}, connects more devices with high bandwidth memory and will allow for greater use of parallelism and alleviate communication boundedness.

\paragraph{Performance measures and scaling laws must be compute and communication optimal }
Total number of Floating Point Operations (FLOPs) is commonly used to guide the development of efficient model architectures and compute-optimal scaling laws \citep{hoffmann2022empirical, tay2023scaling, dehghaniefficiency}. Without properly accounting for communication dynamics, performance measures and scaling laws cannot be extrapolated from small to large-scale. Integrating holistic information about hardware into scaling practice is essential given that collective communication dominates execution time at scale; scaling laws should be both \textit{compute and communication optimal}.

%% file: sections/4_related_work.tex
\section{Related Work}
\paragraph*{Methods for Training at Scale}
While data, tensor and pipeline parallelization and FSDP are among the most common methods for distributed training of large neural networks, other approaches have been developed to the memory limitations and communication overhead of distributed training.

To address GPU memory limitations, numerous solutions have been proposed which: reduce the storage requirements of training workloads; or utilize offloading to lower bandwidth CPU memory.
Activation checkpointing \citep{griewank2000algorithm, chen2016training} reduces peak memory utilization by discarding intermediate activations during the forward pass and recomputing activations for gradient calculation during the backward pass as needed. Strategies that determine optimal schedules for activation recomputation have been developed to manage the trade-off between activation memory and computational costs using hand-designed schedules or constraint solvers \citep{jain2020checkmate,korthikanti2023reducing, yuan2024accelerating}.

Alternatively, activation compression and reconstruction is an alternative approach to alleviate memory pressure to checkpointing \citep{evans2021ac,georgiadis2019accelerating,liu2021exact,liu2022gact}. Both approaches trade off additional computational overhead for reduced memory utilization. Heterogeneous CPU-GPU methods extend the memory sharding approaches introduced by FSDP and ZeRO to offload parameters, gradients, and optimizer states to larger RAM and NVMe memory \citep{rajbhandari2021zero,ren2021zero}. However, these methods incur substantial data transfer costs relying on CPU and PCI-E memory bandwidth orders of magnitude slower than GPU memory.

Communication overhead increases as the number of devices increases, which requires methods to reduce communication load. Hierarchical parallelization strategies such as Hybrid-Sharded Data Parallelism (HSDP, \citeauthor{fbFullySharded}) 
Algorithmic variations of standard minibatch SGD reduce communication volume by performing less frequent parameter updates via federated averaging, such as Diloco, Local SGD, Model Soups, and Branch-Train-Merge \citep{douillard2023diloco,stich2018local, li2022branch,wortsman2022model}. 






\paragraph*{Evaluations of Parallelization Strategies.}
Previous studies empirically evaluating the scaling properties of distributed training strategies for neural networks has largely focused on the interaction of model parallelism with standard data parallelism techniques in the absence of FSDP or ZeRO-3 parallelism \citep{hagemann2023efficient, narayanan2021efficient}. Such studies recommend that total model parallelism be minimized due to the additional communication operations and overhead introduced by model parallelism, which we show does not apply when training with FSDP of Zero-3, alone.

Complementing empirical studies, automatic parallelization strategies and cost models for distributed training have been developed; such as Alpa, Galvatron, and FlexFlow \citep{zheng2022alpa,miao2022galvatron, lu2017flexflow}.  However, these works limit their validation with smaller models and fewer accelerators (up to 64 GPUs) far less than the world sizes we evaluate in our experiments.

\paragraph*{Scaling Properties of Deep Learning.}
Previous work investigating the scaling properties of neural network training has largely studied the effects of varying the data volume, training compute budget, and model architecture \citep{hoffmann2022empirical, kaplan2020scaling, tay2023scaling, porian2024resolving}. These works primarily examine the impact of these factors on the pretraining loss and downstream finetuning performance of the model with respect to the theoretical amount of computational resources allocated (i.e. number of FLOPs).

However, these analyses assume that workload performance scales directly with the amount of computation regardless of the underlying hardware platform and frameworks. In practice, theoretical measures (i.e. FLOPs) are known to be imprecise representations of end-to-end real-world performance (e.g. latency, throughput) due to performance bounds that emerge from management of the computational graph, data transfer, and communication bottlenecks \citep{dehghaniefficiency, fernandez2023framework} -- or as we highlight due to communication boundedness.


%% file: sections/9_conclusion.tex
\section{Conclusion}
    In this work, we examine the effects of hardwdare scaling during the large-scale distributed training of large language models. Specifically, we conduct a comprehensive study of the impact of parallelization strategies, model architectures, and hardware platforms on throughput and energy efficiency during scaling with sharded data parallelism.
    We highlight that while sharded data parallelism is effective at reducing memory utilization when training in smaller regimes, communication boundedness dominates large-scale distributed training and results in reduced hardware utilization.
    
    We show that communication boundedness worsens at scale and with newer hardware generations, and are persistent across model sizes. Additionally, we show that these trends lead to the emergence of viable model parallelism alternatives for distributing deep learning training workloads in contrast to existing recommendations and best practices in regards to training parallelization.  Finally, we show that these trends culminate in significant diminishing returns on training performance with respect to real-world resources of power and throughput.


%% file: sections/8_limitations.tex
\pagebreak

\section{Limitations and Statement of Broader Impact}
In this work, we consider the set of data and model parallelization techniques for distributing training of neural networks -- primarily focusing on the interactions of scale and parallelization in the Fully Sharded Data Parallel Setting as commonly used in the training of many open models. However, there are additional methods for workload parallelization and memory footprint reduction such as activation checkpointing, Hybrid Sharded Data Parallelism, and asynchronous algorithmic methods for optimization. 

In our investigation across computing platforms, we primarily consider variations in the speed of compute (i.e. GPU generation). In future work, we plan to demonstrate the consistency of the observed trends across settings with variable speeds of communication (i.e. varying speed of internode fabric by comparing InfiniBand interconnects with common alternatives such as RDMA over Converged Ethernet, RoCE).

Additionally, our work is focuses on the training of neural networks based on the Transformer neural network architecture and GPU hardware accelerators. Although we expect our findings to be consistent across other model architectures and hardware platforms, we reserve that examination as areas for future work. Likewise, we focus our investigations on GPUs as it is the most commonly used and easily available hardware accelerator. We expect that similar trends and tradeoffs between communication and computation would occur for alternative hardware accelerator architectures such as TPUs, IPUs, etc.; however we leave exploration of these settings for future study.

One of the primary goals of our work is to provide guidance and best practices for researchers and practitioners training large language models in order to reduce the computational, financial, and environmental impact of training. However, in doing so, this may incentivize further growth in training workloads leading to greater energy use and environmental harm from model training (i.e. Jevon's Paradox).



%% file: sections/A_appx.tex
\section{Software, Hardware, and Dataset Details}
\label{appx:exp_details}
Training is conducted in \texttt{bfloat16} precision with a Megatron-inspired framework and further optimizations provided by FlashAttention-2 \citep{daoflashattention} and xFormers \citep{xFormers2022}. For our primary experiments, we trained models using PyTorch 2.3.1 built with CUDA 12.1, with attention implementation provided by XFormers 0.27. We utilize PyTorch FSDPv2 with prefetch of subsequent layers enabled. 

For the A100 and H100 clusters, intra-node GPU communication occurs via fully connected second and third generation NVLink with NVSwitch, respectively. Inter-node communication occurs over an Infiniband fabric with 200 GB/s and 400 GB/s per-node bandwidth, respectively.

In supplementary experiments with V100 GPUs in Appendix \ref{appx:v100}, models are trained in \texttt{fp16} with loss rescaling and CUTLASS \citep{Thakkar_CUTLASS_2023} attention kernels on Volta hardware -- due to limited hardware support on older Volta hardware. Nodes within the V100 cluster consist of 8-GPU setups connected with first-generation NVLink in a Hybrid Cube Mesh (HCM) topology.

We compute the runtime of communication and computation kernels by using PerfettoSQL to query Kineto profiles extracted by the PyTorch profiler, which is built on top of NVidia CUPTI to identify relevant  NCCL and CUDA kernels, respectively.
In Table \ref{tab:gpu_specs}, we provide additional details on the hardware platforms used for running our experiments.

\label{appx:licenses}
The Llama 2 model is used via the Llama Community License and Acceptable Use Policy. Wikipedia and StackExchange data was made available via Creative Commons Attribution-ShareAlike 4.0 International License (CC BY-SA).

\begin{table*}[h!]
    \small
    \vspace{-0.5em}
    \centering
    \begin{tabular}{l|c|c|c}
        \hline & \textbf{V100 \tablefootnote{\hyperlink{https://www.nvidia.com/content/dam/en-zz/Solutions/Data-Center/dgx-1/dgx-1-rhel-datasheet-nvidia-us-808336-r3-web.pdf}{NVIDIA DGX-1 (V100) Whitepaper}}} &
        \textbf{A100} \tablefootnote{\hyperlink{https://www.nvidia.com/content/dam/en-zz/Solutions/Data-Center/nvidia-dgx-a100-datasheet.pdf}{NVIDIA DGX A100 Whitepaper}} &
        \textbf{H100} \tablefootnote{\hyperlink{https://resources.nvidia.com/en-us-dgx-systems/ai-enterprise-dgx?xs=489753}{NVIDIA DGX H100 Whitepaper}} \\ \hline
        {Tensor Core BF16 FLOPS}& 125 TFLOPS& 312 TFLOPS& 990 TFLOPS \\ \hline
        {GPU HBM} & 900 GB/s & 2 TB/s& 3.35 TB/s \\ \hline
        {NVLink (GPU to GPU Comm)} & 300 GB/s & 600 GB/s& 900 GB/s \\ \hline
        {Internode InfiniBand (Node to Node)}& 100 GB/s & 200 GB/s& 400 GB/s \\ \hline
    \end{tabular}
    \caption{
        Nvidia Reported DGX-Node Specifications by Generation.
    }
    \label{tab:gpu_specs}
\end{table*}

\pagebreak
\section{Additional Experiments: Model Parallelism in Alternate Settings}
\label{appx:alternate_mp_strategies}
\begin{figure*}[h]
    \centering
    \begin{subfigure}{0.9\textwidth}
        \includegraphics[width=\textwidth]{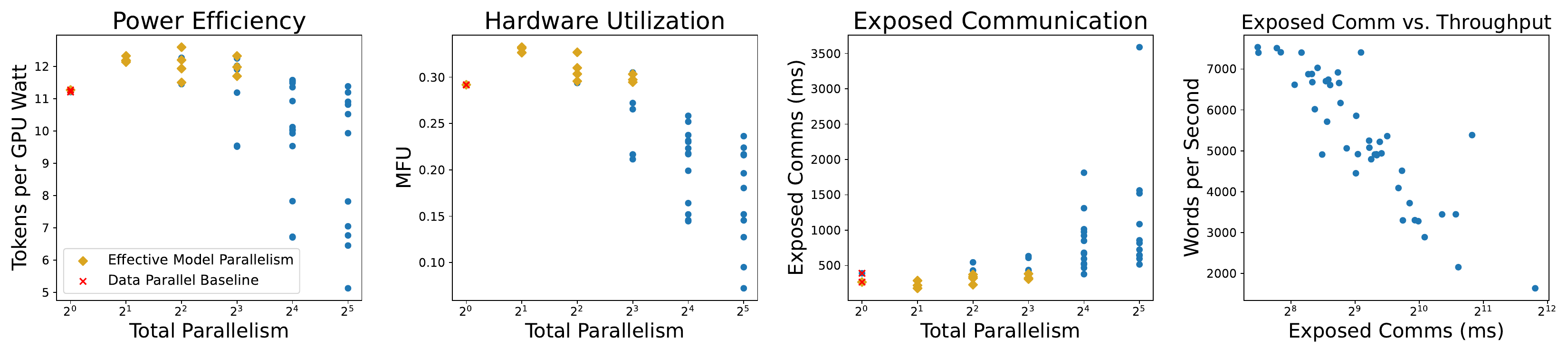}
        \caption{Training Llama-7B with an effective local batch size of 1 on 32 DGX-H100 nodes.}
        \label{fig:mp_bs1}
    \end{subfigure}
    \vspace{1em}
    \begin{subfigure}{0.9\textwidth}
        \includegraphics[width=\textwidth]{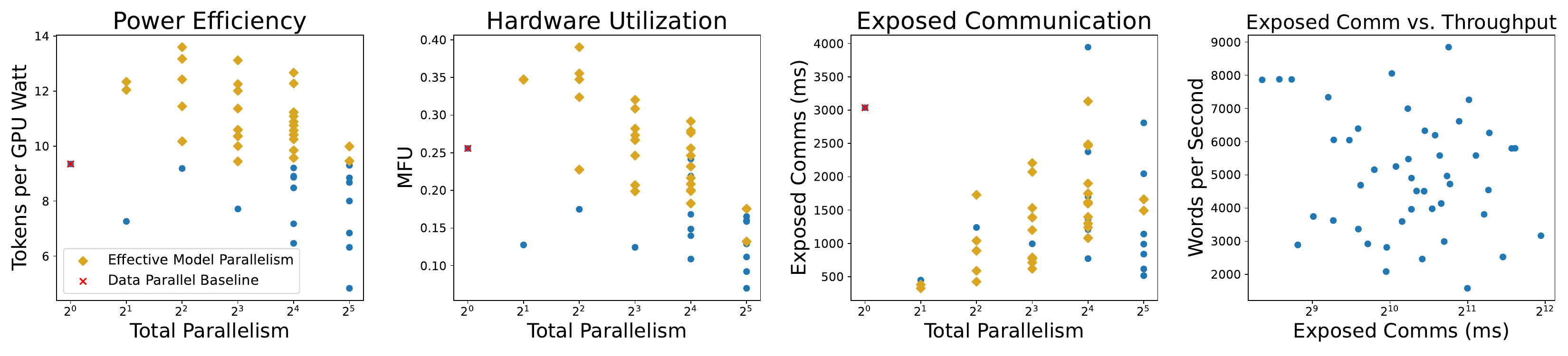}
        \caption{Training Llama-7B with an effective local batch size of 2 on 256 DGX-H100 nodes.}
        \label{fig:mp_2k}
    \end{subfigure}
    \vspace{-2em}
    \caption{
        \small
            In regimes that are low in arithmetic intensity or communication bounded, there exist many viable strategies for model parallelism that: alleviate communication boundedness, increase power efficiency and hardware utilization.
    }
    \label{fig:al}
\end{figure*}

\vspace{-1em}
We extend the experiments from Section \ref{sec:analysis-mp}, in which we examine the effectiveness of model parallelism via Tensor and Pipeline parallelism with evaluations of other hardware settings and computational workloads. 
Here, we consider the effects of model parallelism in settings with lower hardware utilization, due to either: (1) smaller per-device workloads as determined by reduced effective local batch sizes (Figure \ref{fig:mp_bs1}); or (2) larger communication loads from training in a increasingly distributed hardware settings (Figure \ref{fig:mp_2k}). In both regimes, there are a larger number of viable model parallelism strategies.

\vspace{-0.5em}

\vspace{-1em}
\section{Additional Experiments: Fixed Global Batch Size at Pretraining Scale}
\label{appx:fixed_global_batch_size}
\vspace{-0.5em}
We extend the experiments from Section \ref{fig:world_size_scaling}, in which we increase the allocation of hardware accelerators to a fixed computational workload with a constant global batch size -- i.e. increasing the degree of parallelism across more accelerators without increasing the local effective batch size hardware utilization.

\begin{figure}[h]
    \centering
    \begin{subfigure}{\textwidth}
        \centering
        \includegraphics[width=0.9\textwidth]{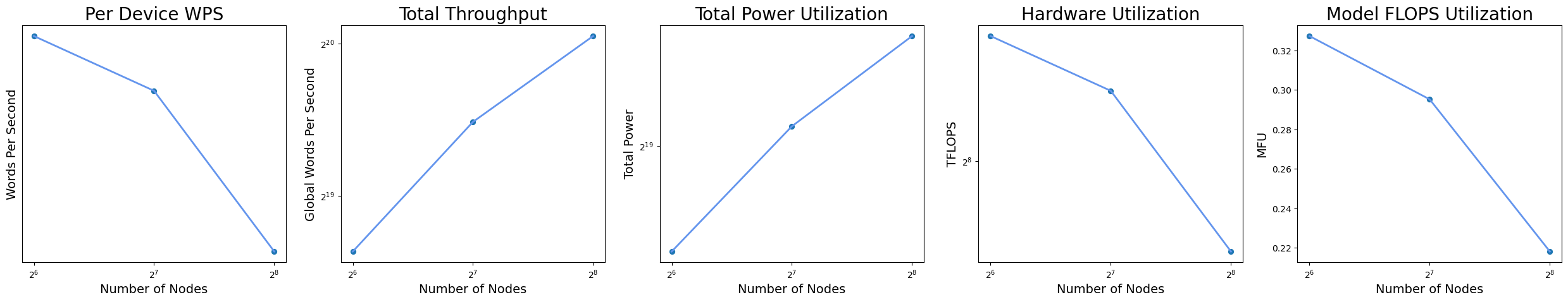}
        \caption{Performance Metrics of Llama-70B Training on 512, 1024, and 2048 GPUs.}
    \end{subfigure}
    \vspace{1em}
    \begin{subfigure}{\textwidth}
        \centering
        \includegraphics[width=0.9\textwidth]{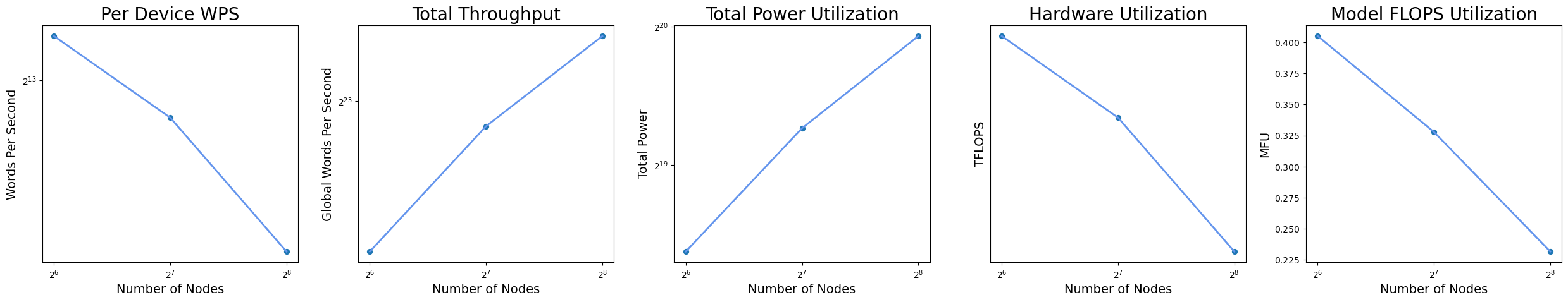}
        \caption{Performance Metrics of Llama-7B Training on 512, 1024, and 2048 GPUs.}
    \end{subfigure}
    \vspace{-2em}
    \caption{
        \small
        At pretraining scale, both Llama-7B and 70B observe regressions in hardware utilization and per-device local throughput as the number of devices is increased for a fixed computational workload.
    }
    \vspace{-1em}
    \label{fig:fixed_batch_pretrain}
\end{figure}

\pagebreak
\section{Additional Experiments: Context Parallelism}
\label{appx:context_parallelism}

\begin{wrapfigure}{l}{0.55\textwidth}
    \vspace{-2.5em}
    \centering
    \includegraphics[width=0.55\textwidth]{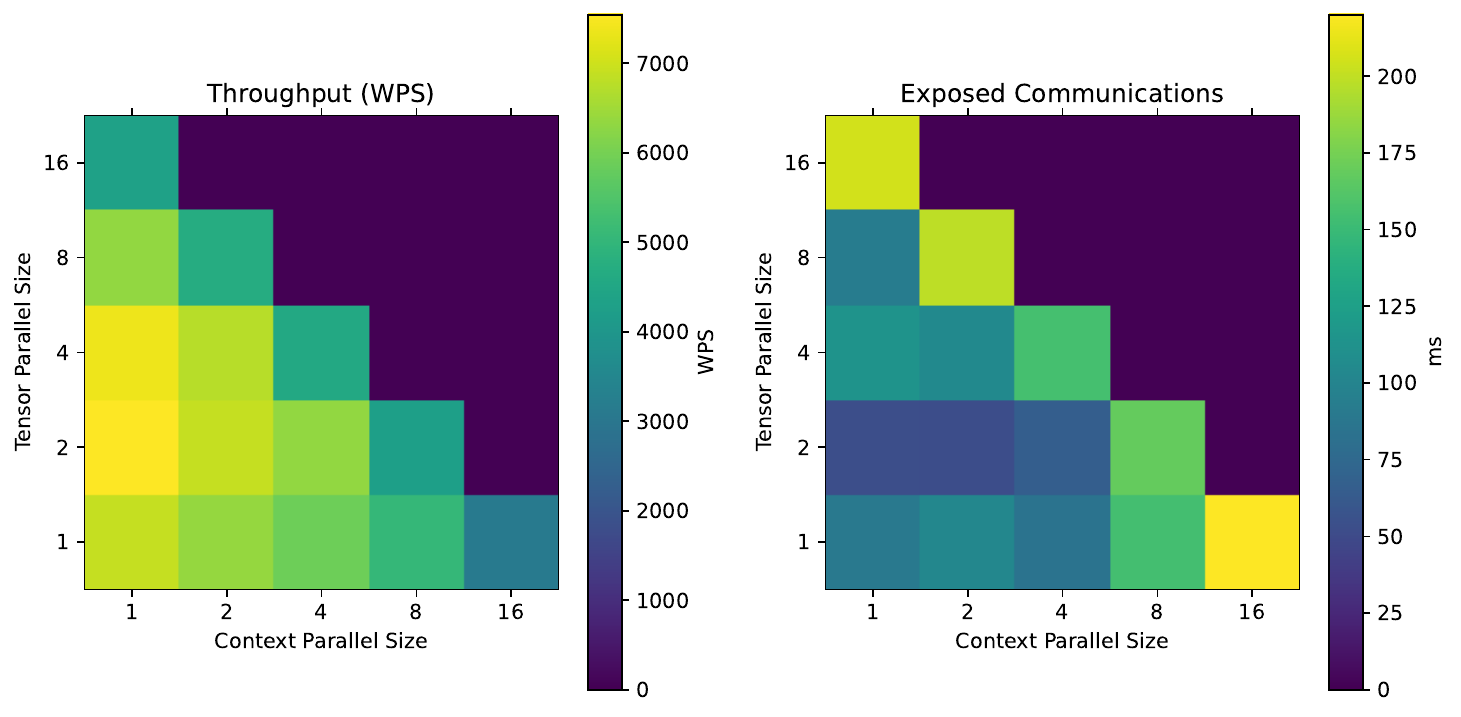}
    \caption{Effectiveness of Context Parallelism in training a Llama-7B model on 4k sequence length with H100 GPUs.}
    \label{fig:context_parallel}
    \vspace{-3em}
\end{wrapfigure}

We extend the results Section \ref{sec:analysis-mp}, to examine an additional form of parallelism, context parallelization \cite{dubey2024llama}. We use the context parallelization implementation provided by Nvidia's TransformerEngine. 
As context parallel is primarily used for very long contexts in Llama-3.1 with sequence lengths of 131,072, we find that context parallelism is a sub-optimal alternative to standard tensor parallelism for relatively common shorter sequence lengths of 4096.

\vspace{2em}

\section{Additional Experiments: V100 Hardware}
\label{appx:v100}
\begin{wrapfigure}{l}{0.55\textwidth}
    \centering

    \includegraphics[width=\linewidth]{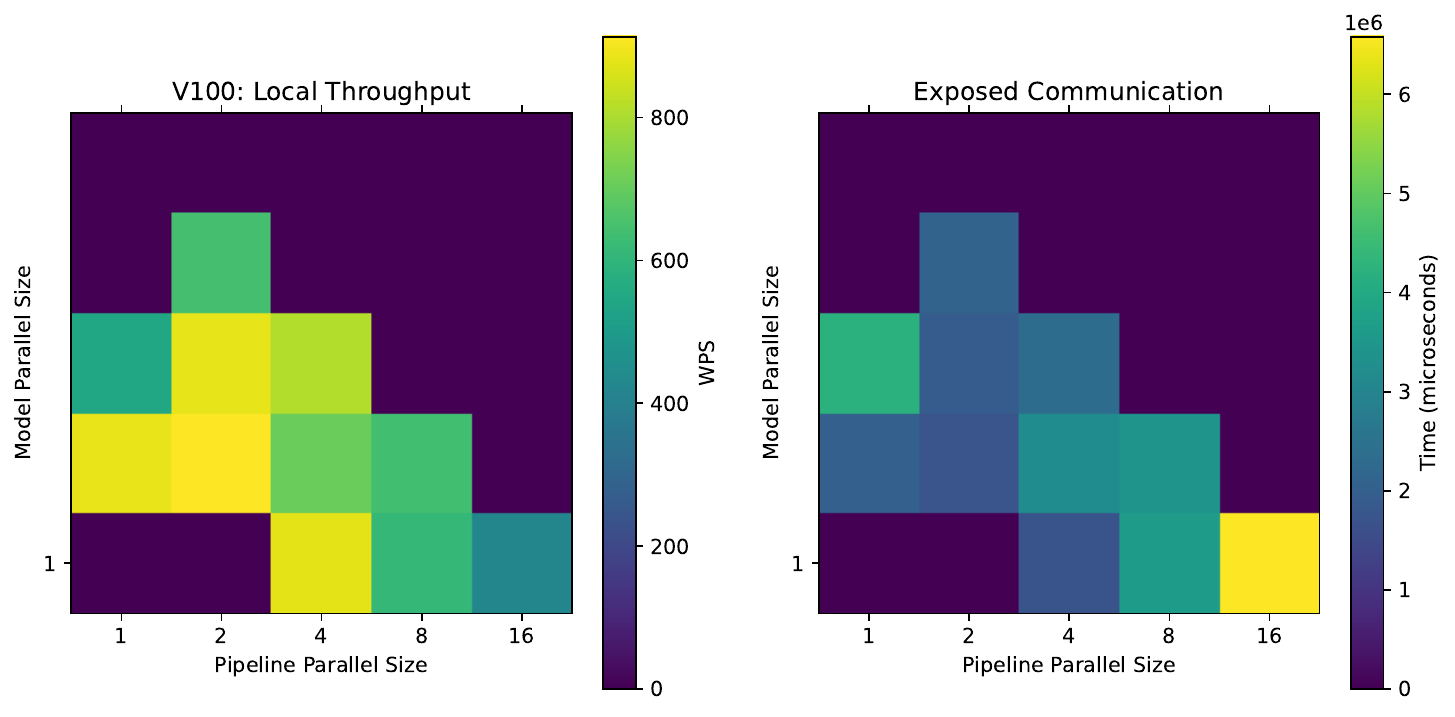}
    \caption{Throughput \& Exposed Communication for Tensor Model Parallelization and Pipeline Parallelism Strategies on V100.}
    \label{fig:v100}
    \vspace{-4em}
\end{wrapfigure}

In addition to our experiments in Section \ref{sec:analysis-mp}, we conduct additional experiments using older V100 GPUs from the Volta architecture training a Llama-7B model with an effective local batch size of 1 on 32 nodes. We observe similar trends in which small degrees of model parallelism improve overall throughput at scale. However, due to lack of optimized kernels (e.g. CUTLASS vs FlashAttention kernels) and Ampere hardware optimizations, we observe that the transition to Ampere A100 GPUs in fact improves overall hardware utilization.

\vspace{2em}
\section{Effects of Scaling on Memory Utilization}
\label{appx:gpu-memory}

\begin{wrapfigure}{l}{0.55\textwidth}
    \centering
    \includegraphics[width=0.6\linewidth]{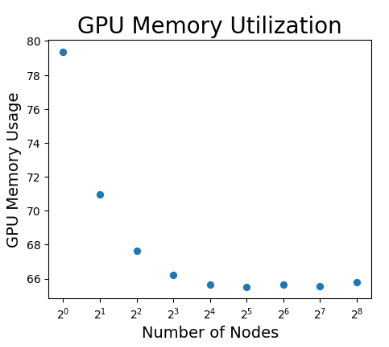}
    \caption{
        Increasing the data parallel group size reduces local per-GPU memory utilization, but reductions diminish with scale.
    }
    \vspace{-6em}
    \label{fig:enter-label}
\end{wrapfigure}

In fully-sharded data parallelism (FSDP), increasing the number of data parallel instances decreases per-GPU memory utilization by sharding parameters and gradients additional data parallel instances. However, memory savings diminish with device world size.

%% file: main.bbl
\begin{thebibliography}{74}
\providecommand{\natexlab}[1]{#1}
\providecommand{\url}[1]{\texttt{#1}}
\expandafter\ifx\csname urlstyle\endcsname\relax
  \providecommand{\doi}[1]{doi: #1}\else
  \providecommand{\doi}{doi: \begingroup \urlstyle{rm}\Url}\fi

\bibitem[Abadi et~al.(2015)Abadi, Agarwal, Barham, Brevdo, Chen, Citro, Corrado, Davis, Dean, Devin, Ghemawat, Goodfellow, Harp, Irving, Isard, Jozefowicz, Jia, Kaiser, Kudlur, Levenberg, Mané, Schuster, Monga, Moore, Murray, Olah, Shlens, Steiner, Sutskever, Talwar, Tucker, Vanhoucke, Vasudevan, Viégas, Vinyals, Warden, Wattenberg, Wicke, Yu, and Zheng]{Abadi_TensorFlow_Large-scale_machine_2015}
Martín Abadi, Ashish Agarwal, Paul Barham, Eugene Brevdo, Zhifeng Chen, Craig Citro, Greg~S. Corrado, Andy Davis, Jeffrey Dean, Matthieu Devin, Sanjay Ghemawat, Ian Goodfellow, Andrew Harp, Geoffrey Irving, Michael Isard, Rafal Jozefowicz, Yangqing Jia, Lukasz Kaiser, Manjunath Kudlur, Josh Levenberg, Dan Mané, Mike Schuster, Rajat Monga, Sherry Moore, Derek Murray, Chris Olah, Jonathon Shlens, Benoit Steiner, Ilya Sutskever, Kunal Talwar, Paul Tucker, Vincent Vanhoucke, Vijay Vasudevan, Fernanda Viégas, Oriol Vinyals, Pete Warden, Martin Wattenberg, Martin Wicke, Yuan Yu, and Xiaoqiang Zheng.
\newblock {TensorFlow, Large-scale machine learning on heterogeneous systems}, November 2015.

\bibitem[{BigScience Workshop}(2022)]{bigscience_workshop_2022}
{BigScience Workshop}.
\newblock {BLOOM} (revision 4ab0472), 2022.
\newblock URL \url{https://huggingface.co/bigscience/bloom}.

\bibitem[Bradbury et~al.(2018)Bradbury, Frostig, Hawkins, Johnson, Leary, Maclaurin, Necula, Paszke, Vander{P}las, Wanderman-{M}ilne, and Zhang]{jax2018github}
James Bradbury, Roy Frostig, Peter Hawkins, Matthew~James Johnson, Chris Leary, Dougal Maclaurin, George Necula, Adam Paszke, Jake Vander{P}las, Skye Wanderman-{M}ilne, and Qiao Zhang.
\newblock {JAX}: composable transformations of {P}ython+{N}um{P}y programs, 2018.
\newblock URL \url{http://github.com/jax-ml/jax}.

\bibitem[Cai et~al.(2021)Cai, Yan, Ma, Wu, Huang, Cheng, Su, and Yu]{cai2021tensoropt}
Zhenkun Cai, Xiao Yan, Kaihao Ma, Yidi Wu, Yuzhen Huang, James Cheng, Teng Su, and Fan Yu.
\newblock Tensoropt: Exploring the tradeoffs in distributed dnn training with auto-parallelism.
\newblock \emph{IEEE Transactions on Parallel and Distributed Systems}, 33\penalty0 (8):\penalty0 1967--1981, 2021.

\bibitem[Chen et~al.(2016)Chen, Xu, Zhang, and Guestrin]{chen2016training}
Tianqi Chen, Bing Xu, Chiyuan Zhang, and Carlos Guestrin.
\newblock Training deep nets with sublinear memory cost.
\newblock \emph{arXiv preprint arXiv:1604.06174}, 2016.

\bibitem[Chowdhery et~al.(2023)Chowdhery, Narang, Devlin, Bosma, Mishra, Roberts, Barham, Chung, Sutton, Gehrmann, et~al.]{chowdhery2023palm}
Aakanksha Chowdhery, Sharan Narang, Jacob Devlin, Maarten Bosma, Gaurav Mishra, Adam Roberts, Paul Barham, Hyung~Won Chung, Charles Sutton, Sebastian Gehrmann, et~al.
\newblock Palm: Scaling language modeling with pathways.
\newblock \emph{Journal of Machine Learning Research}, 24\penalty0 (240):\penalty0 1--113, 2023.

\bibitem[Dao(2024)]{daoflashattention}
Tri Dao.
\newblock Flashattention-2: Faster attention with better parallelism and work partitioning.
\newblock In \emph{The Twelfth International Conference on Learning Representations}, 2024.

\bibitem[Dean et~al.(2012)Dean, Corrado, Monga, Chen, Devin, Mao, Ranzato, Senior, Tucker, Yang, et~al.]{dean2012large}
Jeffrey Dean, Greg Corrado, Rajat Monga, Kai Chen, Matthieu Devin, Mark Mao, Marc'aurelio Ranzato, Andrew Senior, Paul Tucker, Ke~Yang, et~al.
\newblock Large scale distributed deep networks.
\newblock \emph{Advances in neural information processing systems}, 25, 2012.

\bibitem[Dehghani et~al.(2022)Dehghani, Tay, Arnab, Beyer, and Vaswani]{dehghaniefficiency}
Mostafa Dehghani, Yi~Tay, Anurag Arnab, Lucas Beyer, and Ashish Vaswani.
\newblock The efficiency misnomer.
\newblock In \emph{International Conference on Learning Representations}, 2022.

\bibitem[Douillard et~al.(2023)Douillard, Feng, Rusu, Chhaparia, Donchev, Kuncoro, Ranzato, Szlam, and Shen]{douillard2023diloco}
Arthur Douillard, Qixuan Feng, Andrei~A Rusu, Rachita Chhaparia, Yani Donchev, Adhiguna Kuncoro, Marc'Aurelio Ranzato, Arthur Szlam, and Jiajun Shen.
\newblock Diloco: Distributed low-communication training of language models.
\newblock \emph{arXiv preprint arXiv:2311.08105}, 2023.

\bibitem[Dubey et~al.(2024)Dubey, Jauhri, Pandey, Kadian, Al-Dahle, Letman, Mathur, Schelten, Yang, Fan, et~al.]{dubey2024llama}
Abhimanyu Dubey, Abhinav Jauhri, Abhinav Pandey, Abhishek Kadian, Ahmad Al-Dahle, Aiesha Letman, Akhil Mathur, Alan Schelten, Amy Yang, Angela Fan, et~al.
\newblock The llama 3 herd of models.
\newblock \emph{arXiv preprint arXiv:2407.21783}, 2024.

\bibitem[Evans \& Aamodt(2021)Evans and Aamodt]{evans2021ac}
R~David Evans and Tor Aamodt.
\newblock Ac-gc: Lossy activation compression with guaranteed convergence.
\newblock \emph{Advances in Neural Information Processing Systems}, 34:\penalty0 27434--27448, 2021.

\bibitem[Fernandez et~al.(2023)Fernandez, Kahn, Na, Bisk, and Strubell]{fernandez2023framework}
Jared Fernandez, Jacob Kahn, Clara Na, Yonatan Bisk, and Emma Strubell.
\newblock The framework tax: Disparities between inference efficiency in nlp research and deployment.
\newblock In \emph{The 2023 Conference on Empirical Methods in Natural Language Processing}, 2023.

\bibitem[Georgiadis(2019)]{georgiadis2019accelerating}
Georgios Georgiadis.
\newblock Accelerating convolutional neural networks via activation map compression.
\newblock In \emph{Proceedings of the IEEE/CVF Conference on Computer Vision and Pattern Recognition}, pp.\  7085--7095, 2019.

\bibitem[Gholami et~al.(2018)Gholami, Azad, Jin, Keutzer, and Buluc]{ModelBatchDomainParallelism}
Amir Gholami, Ariful Azad, Peter Jin, Kurt Keutzer, and Aydin Buluc.
\newblock Integrated model, batch, and domain parallelism in training neural networks.
\newblock In \emph{Proceedings of the 30th on Symposium on Parallelism in Algorithms and Architectures}, SPAA '18, pp.\  77–86, New York, NY, USA, 2018. Association for Computing Machinery.
\newblock ISBN 9781450357999.
\newblock \doi{10.1145/3210377.3210394}.
\newblock URL \url{https://doi.org/10.1145/3210377.3210394}.

\bibitem[Granite~Team(2024)]{granite2024granite}
IBM Granite~Team.
\newblock Granite 3.0 language models, 2024.

\bibitem[Griewank \& Walther(2000)Griewank and Walther]{griewank2000algorithm}
Andreas Griewank and Andrea Walther.
\newblock Algorithm 799: revolve: an implementation of checkpointing for the reverse or adjoint mode of computational differentiation.
\newblock \emph{ACM Transactions on Mathematical Software (TOMS)}, 26\penalty0 (1):\penalty0 19--45, 2000.

\bibitem[Groeneveld et~al.(2024)Groeneveld, Beltagy, Walsh, Bhagia, Kinney, Tafjord, Jha, Ivison, Magnusson, Wang, et~al.]{groeneveld2024olmo}
Dirk Groeneveld, Iz~Beltagy, Pete Walsh, Akshita Bhagia, Rodney Kinney, Oyvind Tafjord, Ananya~Harsh Jha, Hamish Ivison, Ian Magnusson, Yizhong Wang, et~al.
\newblock Olmo: Accelerating the science of language models.
\newblock \emph{arXiv preprint arXiv:2402.00838}, 2024.

\bibitem[Gustafson(1988)]{gustafson1988reevaluating}
John~L Gustafson.
\newblock Reevaluating amdahl's law.
\newblock \emph{Communications of the ACM}, 31\penalty0 (5):\penalty0 532--533, 1988.

\bibitem[Hagemann et~al.(2023)Hagemann, Weinbach, Dobler, Schall, and de~Melo]{hagemann2023efficient}
Johannes Hagemann, Samuel Weinbach, Konstantin Dobler, Maximilian Schall, and Gerard de~Melo.
\newblock Efficient parallelization layouts for large-scale distributed model training.
\newblock \emph{arXiv preprint arXiv:2311.05610}, 2023.

\bibitem[Harlap et~al.(2018)Harlap, Narayanan, Phanishayee, Seshadri, Devanur, Ganger, and Gibbons]{harlap2018pipedreamfastefficientpipeline}
Aaron Harlap, Deepak Narayanan, Amar Phanishayee, Vivek Seshadri, Nikhil Devanur, Greg Ganger, and Phil Gibbons.
\newblock Pipedream: Fast and efficient pipeline parallel dnn training, 2018.
\newblock URL \url{https://arxiv.org/abs/1806.03377}.

\bibitem[Hennessy \& Patterson(2017)Hennessy and Patterson]{hennessy2017computer}
John~L Hennessy and David~A Patterson.
\newblock \emph{Computer architecture: a quantitative approach}.
\newblock Morgan kaufmann, 2017.

\bibitem[Hoffmann et~al.(2022)Hoffmann, Borgeaud, Mensch, Buchatskaya, Cai, Rutherford, de~Las~Casas, Hendricks, Welbl, Clark, et~al.]{hoffmann2022empirical}
Jordan Hoffmann, Sebastian Borgeaud, Arthur Mensch, Elena Buchatskaya, Trevor Cai, Eliza Rutherford, Diego de~Las~Casas, Lisa~Anne Hendricks, Johannes Welbl, Aidan Clark, et~al.
\newblock An empirical analysis of compute-optimal large language model training.
\newblock \emph{Advances in Neural Information Processing Systems}, 35:\penalty0 30016--30030, 2022.

\bibitem[Huang et~al.(2018)Huang, Cheng, Chen, Lee, Ngiam, Le, and Chen]{Huang2018GPipeET}
Yanping Huang, Yonglong Cheng, Dehao Chen, HyoukJoong Lee, Jiquan Ngiam, Quoc~V. Le, and Z.~Chen.
\newblock Gpipe: Efficient training of giant neural networks using pipeline parallelism.
\newblock In \emph{Neural Information Processing Systems}, 2018.
\newblock URL \url{https://api.semanticscholar.org/CorpusID:53670168}.

\bibitem[Jacobs et~al.(2023)Jacobs, Tanaka, Zhang, Zhang, Song, Rajbhandari, and He]{jacobs2023deepspeed}
Sam~Ade Jacobs, Masahiro Tanaka, Chengming Zhang, Minjia Zhang, Shuaiwen~Leon Song, Samyam Rajbhandari, and Yuxiong He.
\newblock Deepspeed ulysses: System optimizations for enabling training of extreme long sequence transformer models.
\newblock \emph{CoRR}, 2023.

\bibitem[Jain et~al.(2020)Jain, Jain, Nrusimha, Gholami, Abbeel, Gonzalez, Keutzer, and Stoica]{jain2020checkmate}
Paras Jain, Ajay Jain, Aniruddha Nrusimha, Amir Gholami, Pieter Abbeel, Joseph Gonzalez, Kurt Keutzer, and Ion Stoica.
\newblock Checkmate: Breaking the memory wall with optimal tensor rematerialization.
\newblock \emph{Proceedings of Machine Learning and Systems}, 2:\penalty0 497--511, 2020.

\bibitem[Jia et~al.(2019)Jia, Zaharia, and Aiken]{jia2019beyond}
Zhihao Jia, Matei Zaharia, and Alex Aiken.
\newblock Beyond data and model parallelism for deep neural networks.
\newblock \emph{Proceedings of Machine Learning and Systems}, 1:\penalty0 1--13, 2019.

\bibitem[Kaplan et~al.(2020)Kaplan, McCandlish, Henighan, Brown, Chess, Child, Gray, Radford, Wu, and Amodei]{kaplan2020scaling}
Jared Kaplan, Sam McCandlish, Tom Henighan, Tom~B Brown, Benjamin Chess, Rewon Child, Scott Gray, Alec Radford, Jeffrey Wu, and Dario Amodei.
\newblock Scaling laws for neural language models.
\newblock \emph{arXiv preprint arXiv:2001.08361}, 2020.

\bibitem[Kingma \& Ba(2015)Kingma and Ba]{KingBa15}
Diederik Kingma and Jimmy Ba.
\newblock Adam: A method for stochastic optimization.
\newblock In \emph{International Conference on Learning Representations (ICLR)}, San Diega, CA, USA, 2015.

\bibitem[Korthikanti et~al.(2023)Korthikanti, Casper, Lym, McAfee, Andersch, Shoeybi, and Catanzaro]{korthikanti2023reducing}
Vijay~Anand Korthikanti, Jared Casper, Sangkug Lym, Lawrence McAfee, Michael Andersch, Mohammad Shoeybi, and Bryan Catanzaro.
\newblock Reducing activation recomputation in large transformer models.
\newblock \emph{Proceedings of Machine Learning and Systems}, 5:\penalty0 341--353, 2023.

\bibitem[Lamy-Poirier(2023)]{lamy2023breadth}
Joel Lamy-Poirier.
\newblock Breadth-first pipeline parallelism.
\newblock \emph{Proceedings of Machine Learning and Systems}, 5:\penalty0 48--67, 2023.

\bibitem[Lefaudeux et~al.(2022)Lefaudeux, Massa, Liskovich, Xiong, Caggiano, Naren, Xu, Hu, Tintore, Zhang, Labatut, Haziza, Wehrstedt, Reizenstein, and Sizov]{xFormers2022}
Benjamin Lefaudeux, Francisco Massa, Diana Liskovich, Wenhan Xiong, Vittorio Caggiano, Sean Naren, Min Xu, Jieru Hu, Marta Tintore, Susan Zhang, Patrick Labatut, Daniel Haziza, Luca Wehrstedt, Jeremy Reizenstein, and Grigory Sizov.
\newblock xformers: A modular and hackable transformer modelling library.
\newblock \url{https://github.com/facebookresearch/xformers}, 2022.

\bibitem[Lepikhin et~al.(2020)Lepikhin, Lee, Xu, Chen, Firat, Huang, Krikun, Shazeer, and Chen]{lepikhin2020gshard}
Dmitry Lepikhin, HyoukJoong Lee, Yuanzhong Xu, Dehao Chen, Orhan Firat, Yanping Huang, Maxim Krikun, Noam Shazeer, and Zhifeng Chen.
\newblock Gshard: Scaling giant models with conditional computation and automatic sharding.
\newblock In \emph{International Conference on Learning Representations}, 2020.

\bibitem[Li et~al.(2022)Li, Gururangan, Dettmers, Lewis, Althoff, Smith, and Zettlemoyer]{li2022branch}
Margaret Li, Suchin Gururangan, Tim Dettmers, Mike Lewis, Tim Althoff, Noah~A Smith, and Luke Zettlemoyer.
\newblock Branch-train-merge: Embarrassingly parallel training of expert language models.
\newblock In \emph{First Workshop on Interpolation Regularizers and Beyond at NeurIPS 2022}, 2022.

\bibitem[Li et~al.(2020)Li, Zhao, Varma, Salpekar, Noordhuis, Li, Paszke, Smith, Vaughan, Damania, and Chintala]{Li2020PyTorchD}
Shen Li, Yanli Zhao, Rohan Varma, Omkar Salpekar, Pieter Noordhuis, Teng Li, Adam Paszke, Jeff Smith, Brian Vaughan, Pritam Damania, and Soumith Chintala.
\newblock Pytorch distributed.
\newblock \emph{Proceedings of the VLDB Endowment}, 13:\penalty0 3005 -- 3018, 2020.
\newblock URL \url{https://api.semanticscholar.org/CorpusID:220250008}.

\bibitem[Li et~al.(2023)Li, Xue, Baranwal, Li, and You]{li2023sequence}
Shenggui Li, Fuzhao Xue, Chaitanya Baranwal, Yongbin Li, and Yang You.
\newblock Sequence parallelism: Long sequence training from system perspective.
\newblock In \emph{The 61st Annual Meeting Of The Association For Computational Linguistics}, 2023.

\bibitem[Li \& Hoefler(2021)Li and Hoefler]{li2021chimera}
Shigang Li and Torsten Hoefler.
\newblock Chimera: efficiently training large-scale neural networks with bidirectional pipelines.
\newblock In \emph{Proceedings of the International Conference for High Performance Computing, Networking, Storage and Analysis}, pp.\  1--14, 2021.

\bibitem[Li et~al.(2021)Li, Zhuang, Guo, Zhuo, Zhang, Song, and Stoica]{li2021terapipe}
Zhuohan Li, Siyuan Zhuang, Shiyuan Guo, Danyang Zhuo, Hao Zhang, Dawn Song, and Ion Stoica.
\newblock Terapipe: Token-level pipeline parallelism for training large-scale language models.
\newblock In \emph{International Conference on Machine Learning}, pp.\  6543--6552. PMLR, 2021.

\bibitem[Liu et~al.(2024)Liu, Zaharia, and Abbeel]{liuringattention}
Hao Liu, Matei Zaharia, and Pieter Abbeel.
\newblock Ringattention with blockwise transformers for near-infinite context.
\newblock In \emph{The Twelfth International Conference on Learning Representations}, 2024.

\bibitem[Liu et~al.(2022)Liu, Zheng, Wang, Cen, Chen, Han, Chen, Liu, Tang, Gonzalez, et~al.]{liu2022gact}
Xiaoxuan Liu, Lianmin Zheng, Dequan Wang, Yukuo Cen, Weize Chen, Xu~Han, Jianfei Chen, Zhiyuan Liu, Jie Tang, Joey Gonzalez, et~al.
\newblock Gact: Activation compressed training for generic network architectures.
\newblock In \emph{International Conference on Machine Learning}, pp.\  14139--14152. PMLR, 2022.

\bibitem[Liu et~al.(2021)Liu, Zhou, Yang, Li, Chen, and Hu]{liu2021exact}
Zirui Liu, Kaixiong Zhou, Fan Yang, Li~Li, Rui Chen, and Xia Hu.
\newblock Exact: Scalable graph neural networks training via extreme activation compression.
\newblock In \emph{International Conference on Learning Representations}, 2021.

\bibitem[Loshchilov \& Hutter(2019)Loshchilov and Hutter]{loshchilov2018decoupled}
Ilya Loshchilov and Frank Hutter.
\newblock Decoupled weight decay regularization.
\newblock In \emph{International Conference on Learning Representations}, 2019.
\newblock URL \url{https://openreview.net/forum?id=Bkg6RiCqY7}.

\bibitem[Lu et~al.(2017)Lu, Yan, Li, Gong, Han, and Li]{lu2017flexflow}
Wenyan Lu, Guihai Yan, Jiajun Li, Shijun Gong, Yinhe Han, and Xiaowei Li.
\newblock Flexflow: A flexible dataflow accelerator architecture for convolutional neural networks.
\newblock In \emph{2017 IEEE international symposium on high performance computer architecture (HPCA)}, pp.\  553--564. IEEE, 2017.

\bibitem[Luccioni \& Hernandez-Garcia(2023)Luccioni and Hernandez-Garcia]{luccioni2023countingcarbonsurveyfactors}
Alexandra~Sasha Luccioni and Alex Hernandez-Garcia.
\newblock Counting carbon: A survey of factors influencing the emissions of machine learning, 2023.
\newblock URL \url{https://arxiv.org/abs/2302.08476}.

\bibitem[Luccioni et~al.(2024)Luccioni, Jernite, and Strubell]{luccioni2024power}
Sasha Luccioni, Yacine Jernite, and Emma Strubell.
\newblock Power hungry processing: Watts driving the cost of ai deployment?
\newblock In \emph{The 2024 ACM Conference on Fairness, Accountability, and Transparency}, pp.\  85--99, 2024.

\bibitem[Mehta et~al.(2024)Mehta, Sekhavat, Cao, Horton, Jin, Sun, Mirzadeh, Najibi, Belenko, Zatloukal, et~al.]{mehta2024openelm}
Sachin Mehta, Mohammad~Hossein Sekhavat, Qingqing Cao, Maxwell Horton, Yanzi Jin, Chenfan Sun, Seyed~Iman Mirzadeh, Mahyar Najibi, Dmitry Belenko, Peter Zatloukal, et~al.
\newblock Openelm: An efficient language model family with open training and inference framework.
\newblock In \emph{Workshop on Efficient Systems for Foundation Models II@ ICML2024}, 2024.

\bibitem[Miao et~al.(2022)Miao, Wang, Jiang, Shi, Nie, Zhang, and Cui]{miao2022galvatron}
Xupeng Miao, Yujie Wang, Youhe Jiang, Chunan Shi, Xiaonan Nie, Hailin Zhang, and Bin Cui.
\newblock Galvatron: Efficient transformer training over multiple gpus using automatic parallelism.
\newblock \emph{Proceedings of the VLDB Endowment}, 16\penalty0 (3):\penalty0 470--479, 2022.

\bibitem[Narayanan et~al.(2019)Narayanan, Harlap, Phanishayee, Seshadri, Devanur, Ganger, Gibbons, and Zaharia]{Narayanan2019PipeDreamGP}
Deepak Narayanan, Aaron Harlap, Amar Phanishayee, Vivek Seshadri, Nikhil~R. Devanur, Gregory~R. Ganger, Phillip~B. Gibbons, and Matei~A. Zaharia.
\newblock Pipedream: generalized pipeline parallelism for dnn training.
\newblock \emph{Proceedings of the 27th ACM Symposium on Operating Systems Principles}, 2019.
\newblock URL \url{https://api.semanticscholar.org/CorpusID:202488191}.

\bibitem[Narayanan et~al.(2021)Narayanan, Shoeybi, Casper, LeGresley, Patwary, Korthikanti, Vainbrand, Kashinkunti, Bernauer, Catanzaro, et~al.]{narayanan2021efficient}
Deepak Narayanan, Mohammad Shoeybi, Jared Casper, Patrick LeGresley, Mostofa Patwary, Vijay Korthikanti, Dmitri Vainbrand, Prethvi Kashinkunti, Julie Bernauer, Bryan Catanzaro, et~al.
\newblock Efficient large-scale language model training on gpu clusters using megatron-lm.
\newblock In \emph{Proceedings of the International Conference for High Performance Computing, Networking, Storage and Analysis}, pp.\  1--15, 2021.

\bibitem[Ott et~al.(2024)Ott, Shleifer, Xu, Goyal, Duval, and Caggiano]{fbFullySharded}
Myle Ott, Sam Shleifer, Min Xu, Priya Goyal, Quentin Duval, and Vittorio Caggiano.
\newblock Fully sharded data parallel: faster ai training with fewer gpus --- engineering.fb.com.
\newblock \url{https://engineering.fb.com/2021/07/15/open-source/fsdp/}, 2024.
\newblock [Accessed 30-09-2024].

\bibitem[Pal et~al.(2019)Pal, Ebrahimi, Zulfiqar, Fu, Zhang, Migacz, Nellans, and Gupta]{pal2019optimizing}
Saptadeep Pal, Eiman Ebrahimi, Arslan Zulfiqar, Yaosheng Fu, Victor Zhang, Szymon Migacz, David Nellans, and Puneet Gupta.
\newblock Optimizing multi-gpu parallelization strategies for deep learning training.
\newblock \emph{Ieee Micro}, 39\penalty0 (5):\penalty0 91--101, 2019.

\bibitem[Paszke et~al.(2019)Paszke, Gross, Massa, Lerer, Bradbury, Chanan, Killeen, Lin, Gimelshein, Antiga, et~al.]{paszke2019pytorch}
Adam Paszke, Sam Gross, Francisco Massa, Adam Lerer, James Bradbury, Gregory Chanan, Trevor Killeen, Zeming Lin, Natalia Gimelshein, Luca Antiga, et~al.
\newblock Pytorch: An imperative style, high-performance deep learning library.
\newblock \emph{Advances in neural information processing systems}, 32, 2019.

\bibitem[Porian et~al.(2024)Porian, Wortsman, Jitsev, Schmidt, and Carmon]{porian2024resolving}
Tomer Porian, Mitchell Wortsman, Jenia Jitsev, Ludwig Schmidt, and Yair Carmon.
\newblock Resolving discrepancies in compute-optimal scaling of language models.
\newblock \emph{arXiv preprint arXiv:2406.19146}, 2024.

\bibitem[Qi et~al.(2017)Qi, Sparks, and Talwalkar]{qi2017paleo}
Hang Qi, Evan~R Sparks, and Ameet Talwalkar.
\newblock Paleo: A performance model for deep neural networks.
\newblock In \emph{International Conference on Learning Representations}, 2017.

\bibitem[Rajbhandari et~al.(2020)Rajbhandari, Rasley, Ruwase, and He]{rajbhandari2020zero}
Samyam Rajbhandari, Jeff Rasley, Olatunji Ruwase, and Yuxiong He.
\newblock Zero: Memory optimizations toward training trillion parameter models.
\newblock In \emph{SC20: International Conference for High Performance Computing, Networking, Storage and Analysis}, pp.\  1--16. IEEE, 2020.

\bibitem[Rajbhandari et~al.(2021)Rajbhandari, Ruwase, Rasley, Smith, and He]{rajbhandari2021zero}
Samyam Rajbhandari, Olatunji Ruwase, Jeff Rasley, Shaden Smith, and Yuxiong He.
\newblock Zero-infinity: Breaking the gpu memory wall for extreme scale deep learning.
\newblock In \emph{Proceedings of the international conference for high performance computing, networking, storage and analysis}, pp.\  1--14, 2021.

\bibitem[Rasley et~al.(2020)Rasley, Rajbhandari, Ruwase, and He]{rasley2020deepspeed}
Jeff Rasley, Samyam Rajbhandari, Olatunji Ruwase, and Yuxiong He.
\newblock Deepspeed: System optimizations enable training deep learning models with over 100 billion parameters.
\newblock In \emph{Proceedings of the 26th ACM SIGKDD International Conference on Knowledge Discovery \& Data Mining}, pp.\  3505--3506, 2020.

\bibitem[Ren et~al.(2021)Ren, Rajbhandari, Aminabadi, Ruwase, Yang, Zhang, Li, and He]{ren2021zero}
Jie Ren, Samyam Rajbhandari, Reza~Yazdani Aminabadi, Olatunji Ruwase, Shuangyan Yang, Minjia Zhang, Dong Li, and Yuxiong He.
\newblock $\{$Zero-offload$\}$: Democratizing $\{$billion-scale$\}$ model training.
\newblock In \emph{2021 USENIX Annual Technical Conference (USENIX ATC 21)}, pp.\  551--564, 2021.

\bibitem[Ryabinin et~al.(2023)Ryabinin, Dettmers, Diskin, and Borzunov]{ryabinin2023swarm}
Max Ryabinin, Tim Dettmers, Michael Diskin, and Alexander Borzunov.
\newblock Swarm parallelism: Training large models can be surprisingly communication-efficient.
\newblock In \emph{International Conference on Machine Learning}, pp.\  29416--29440. PMLR, 2023.

\bibitem[Schwartz et~al.(2020)Schwartz, Dodge, Smith, and Etzioni]{schwartz2020green}
Roy Schwartz, Jesse Dodge, Noah~A Smith, and Oren Etzioni.
\newblock Green ai.
\newblock \emph{Communications of the ACM}, 63\penalty0 (12):\penalty0 54--63, 2020.

\bibitem[Shazeer et~al.(2017)Shazeer, Mirhoseini, Maziarz, Davis, Le, Hinton, and Dean]{shazeer2017outrageously}
Noam Shazeer, Azalia Mirhoseini, Krzysztof Maziarz, Andy Davis, Quoc Le, Geoffrey Hinton, and Jeff Dean.
\newblock Outrageously large neural networks: The sparsely-gated mixture-of-experts layer.
\newblock \emph{arXiv preprint arXiv:1701.06538}, 2017.

\bibitem[Shazeer et~al.(2018)Shazeer, Cheng, Parmar, Tran, Vaswani, Koanantakool, Hawkins, Lee, Hong, Young, Sepassi, and Hechtman]{shazeer2018meshtensorflowdeeplearningsupercomputers}
Noam Shazeer, Youlong Cheng, Niki Parmar, Dustin Tran, Ashish Vaswani, Penporn Koanantakool, Peter Hawkins, HyoukJoong Lee, Mingsheng Hong, Cliff Young, Ryan Sepassi, and Blake Hechtman.
\newblock Mesh-tensorflow: Deep learning for supercomputers, 2018.
\newblock URL \url{https://arxiv.org/abs/1811.02084}.

\bibitem[Shoeybi et~al.(2019)Shoeybi, Patwary, Puri, LeGresley, Casper, and Catanzaro]{Shoeybi2019MegatronLMTM}
Mohammad Shoeybi, Mostofa Patwary, Raul Puri, Patrick LeGresley, Jared Casper, and Bryan Catanzaro.
\newblock Megatron-lm: Training multi-billion parameter language models using model parallelism.
\newblock \emph{ArXiv}, abs/1909.08053, 2019.
\newblock URL \url{https://api.semanticscholar.org/CorpusID:202660670}.

\bibitem[Stich(2018)]{stich2018local}
Sebastian~U Stich.
\newblock Local sgd converges fast and communicates little.
\newblock \emph{arXiv preprint arXiv:1805.09767}, 2018.

\bibitem[Strubell et~al.(2019)Strubell, Ganesh, and McCallum]{strubell-etal-2019-energy}
Emma Strubell, Ananya Ganesh, and Andrew McCallum.
\newblock Energy and policy considerations for deep learning in {NLP}.
\newblock In Anna Korhonen, David Traum, and Llu{\'\i}s M{\`a}rquez (eds.), \emph{Proceedings of the 57th Annual Meeting of the Association for Computational Linguistics}, pp.\  3645--3650, Florence, Italy, July 2019. Association for Computational Linguistics.
\newblock \doi{10.18653/v1/P19-1355}.
\newblock URL \url{https://aclanthology.org/P19-1355}.

\bibitem[Tay et~al.(2023)Tay, Dehghani, Abnar, Chung, Fedus, Rao, Narang, Tran, Yogatama, and Metzler]{tay2023scaling}
Yi~Tay, Mostafa Dehghani, Samira Abnar, Hyung~Won Chung, William Fedus, Jinfeng Rao, Sharan Narang, Vinh~Q. Tran, Dani Yogatama, and Donald Metzler.
\newblock Scaling laws vs model architectures: How does inductive bias influence scaling?
\newblock In \emph{The 2023 Conference on Empirical Methods in Natural Language Processing}, 2023.
\newblock URL \url{https://openreview.net/forum?id=E9dH0BP5VW}.

\bibitem[Team(2023)]{MosaicML2023Introducing}
MosaicML~NLP Team.
\newblock Introducing mpt-7b: A new standard for open-source, commercially usable llms, 2023.
\newblock URL \url{www.mosaicml.com/blog/mpt-7b}.
\newblock Accessed: 2023-05-05.

\bibitem[Thakkar et~al.(2023)Thakkar, Ramani, Cecka, Shivam, Lu, Yan, Kosaian, Hoemmen, Wu, Kerr, Nicely, Merrill, Blasig, Qiao, Majcher, Springer, Hohnerbach, Wang, and Gupta]{Thakkar_CUTLASS_2023}
Vijay Thakkar, Pradeep Ramani, Cris Cecka, Aniket Shivam, Honghao Lu, Ethan Yan, Jack Kosaian, Mark Hoemmen, Haicheng Wu, Andrew Kerr, Matt Nicely, Duane Merrill, Dustyn Blasig, Fengqi Qiao, Piotr Majcher, Paul Springer, Markus Hohnerbach, Jin Wang, and Manish Gupta.
\newblock {CUTLASS}, January 2023.
\newblock URL \url{https://github.com/NVIDIA/cutlass}.

\bibitem[Touvron et~al.(2023)Touvron, Martin, Stone, Albert, Almahairi, Babaei, Bashlykov, Batra, Bhargava, Bhosale, et~al.]{touvron2023llama}
Hugo Touvron, Louis Martin, Kevin Stone, Peter Albert, Amjad Almahairi, Yasmine Babaei, Nikolay Bashlykov, Soumya Batra, Prajjwal Bhargava, Shruti Bhosale, et~al.
\newblock Llama 2: Open foundation and fine-tuned chat models.
\newblock \emph{arXiv preprint arXiv:2307.09288}, 2023.

\bibitem[Wortsman et~al.(2022)Wortsman, Ilharco, Gadre, Roelofs, Gontijo-Lopes, Morcos, Namkoong, Farhadi, Carmon, Kornblith, et~al.]{wortsman2022model}
Mitchell Wortsman, Gabriel Ilharco, Samir~Ya Gadre, Rebecca Roelofs, Raphael Gontijo-Lopes, Ari~S Morcos, Hongseok Namkoong, Ali Farhadi, Yair Carmon, Simon Kornblith, et~al.
\newblock Model soups: averaging weights of multiple fine-tuned models improves accuracy without increasing inference time.
\newblock In \emph{International conference on machine learning}, pp.\  23965--23998. PMLR, 2022.

\bibitem[Yang et~al.(2024)Yang, Yang, Ibrahim, Xie, Tang, Sizov, Reizenstein, Park, and Huang]{yang2024context}
Amy Yang, Jingyi Yang, Aya Ibrahim, Xinfeng Xie, Bangsheng Tang, Grigory Sizov, Jeremy Reizenstein, Jongsoo Park, and Jianyu Huang.
\newblock Context parallelism for scalable million-token inference.
\newblock \emph{arXiv preprint arXiv:2411.01783}, 2024.

\bibitem[Yuan et~al.(2024)Yuan, Liu, Ye, Zhang, Tan, Chen, Song, and Zhang]{yuan2024accelerating}
Tailing Yuan, Yuliang Liu, Xucheng Ye, Shenglong Zhang, Jianchao Tan, Bin Chen, Chengru Song, and Di~Zhang.
\newblock Accelerating the training of large language models using efficient activation rematerialization and optimal hybrid parallelism.
\newblock In \emph{2024 USENIX Annual Technical Conference (USENIX ATC 24)}, pp.\  545--561, 2024.

\bibitem[Zhao et~al.(2023)Zhao, Gu, Varma, Luo, chin Huang, Xu, Wright, Shojanazeri, Ott, Shleifer, Desmaison, Balioglu, Nguyen, Chauhan, Hao, and Li]{Zhao2023PyTorchFE}
Yanli Zhao, Andrew Gu, Rohan Varma, Liangchen Luo, Chien chin Huang, Min Xu, Less Wright, Hamid Shojanazeri, Myle Ott, Sam Shleifer, Alban Desmaison, Can Balioglu, Bernard Nguyen, Geeta Chauhan, Yuchen Hao, and Shen Li.
\newblock Pytorch fsdp: Experiences on scaling fully sharded data parallel.
\newblock \emph{Proc. VLDB Endow.}, 16:\penalty0 3848--3860, 2023.
\newblock URL \url{https://api.semanticscholar.org/CorpusID:258297871}.

\bibitem[Zheng et~al.(2022)Zheng, Li, Zhang, Zhuang, Chen, Huang, Wang, Xu, Zhuo, Xing, et~al.]{zheng2022alpa}
Lianmin Zheng, Zhuohan Li, Hao Zhang, Yonghao Zhuang, Zhifeng Chen, Yanping Huang, Yida Wang, Yuanzhong Xu, Danyang Zhuo, Eric~P Xing, et~al.
\newblock Alpa: Automating inter-and $\{$Intra-Operator$\}$ parallelism for distributed deep learning.
\newblock In \emph{16th USENIX Symposium on Operating Systems Design and Implementation (OSDI 22)}, pp.\  559--578, 2022.

\end{thebibliography}
